\documentclass[a4paper,fleqn]{cas-dc}

\usepackage[numbers]{natbib}

\usepackage{amsmath, multirow, booktabs, cleveref, float, wasysym}
\def\tsc#1{\csdef{#1}{\textsc{\lowercase{#1}}\xspace}}
\tsc{WGM}
\tsc{QE}
\tsc{EP}
\tsc{PMS}
\tsc{BEC}
\tsc{DE}

\begin{document}
\let\WriteBookmarks\relax
\def\floatpagepagefraction{1}
\def\textpagefraction{.001}
\shorttitle{}    

\shortauthors{Cho et al.} 

\title [mode = title]{BHaRNet: Reliability-Aware Body-Hand Modality Expertized Networks for Fine-grained Skeleton Action Recognition}                        
\tnotemark[1,2]



\author[1]{Seungyeon Cho}

\cormark[1]

\fnmark[1]

\ead{vinny@kaist.ac.kr}

\ead[url]{https://github.com/VinnyCSY/BHaRNet}

\credit{Conceptualization, Methodology, Software, Validation, Formal analysis, Investigation, Data curation, Writing – original draft, Visualization, Project administration}

\affiliation[1]{organization={School of Computing, KAIST},
            city={Daejeon},
            postcode={34141}, 
            country={Republic of Korea}}

\author[1]{Tae-Kyun Kim}
\cormark[2]
\fnmark[2]

\ead{kimtaekyun@kaist.ac.kr}

\ead[url]{https://sites.google.com/view/tkkim/home}

\credit{Supervision, Resources, Funding acquisition, Writing – review \& editing}

\cortext[cor1]{Corresponding author at: Room 2312, Building E3-2, KAIST, 291 Daehak-ro, Yuseong-gu, Daejeon 34141, Republic of Korea. Phone number: +82) 10-4898-1126.}
\cortext[cor2]{Principal corresponding author}

\nonumnote{}

\begin{abstract}
Skeleton-based human action recognition (HAR) has achieved remarkable progress with graph-based architectures. 
However, most existing methods remain body-centric, focusing on large-scale motions while neglecting subtle hand articulations that are crucial for fine-grained recognition. 
This work presents a probabilistic dual-stream framework that unifies reliability modeling and multi-modal integration, generalizing expertized learning under uncertainty across both intra-skeleton and cross-modal domains. 
The framework comprises three key components: 
(1) a calibration-free preprocessing pipeline that removes canonical-space transformations and learns directly from native coordinates; 
(2) a probabilistic Noisy-OR fusion that stabilizes reliability-aware dual-stream learning without requiring explicit confidence supervision; and 
(3) an intra- to cross-modal ensemble that couples four skeleton modalities (Joint, Bone, Joint Motion, and Bone Motion) to RGB representations, bridging structural and visual motion cues in a unified cross-modal formulation. 
Comprehensive evaluations across multiple benchmarks (NTU RGB+D~60/120, PKU-MMD, N-UCLA) and a newly defined hand-centric benchmark exhibit consistent improvements and robustness under noisy and heterogeneous conditions.
\end{abstract}

\begin{graphicalabstract}
\includegraphics[width=\linewidth]{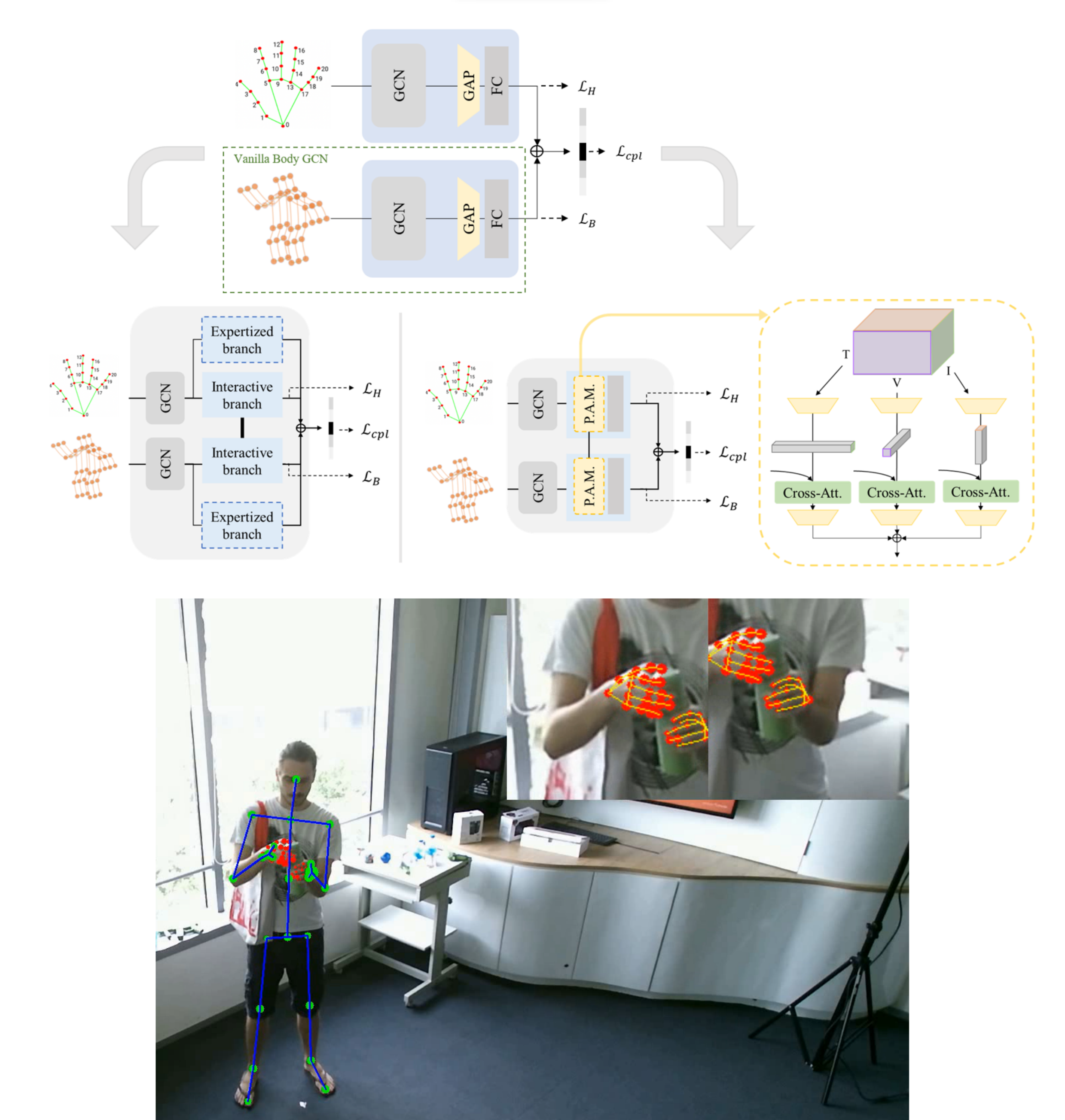}
\end{graphicalabstract}

\begin{highlights}
\item Probabilistic dual-stream body–hand framework with calibration-free skeleton learning and reliability-aware fusion.
\item Noisy-OR–based fusion loss that models asymmetric body–hand reliability and stabilizes dual-stream learning under noisy keypoints.
\item Unified intra- to cross-modal ensemble that extends joint, bone, and motion cues to RGB for efficient skeleton–RGB action recognition.
\end{highlights}


\maketitle
\section{Introduction}

Human action recognition (HAR) is a long-standing challenge in computer vision, with applications in human–computer interaction, video understanding, and behavioral analysis~\cite{wen2023interactive,noisylabels}. 
Among various sensing modalities—RGB, depth, and skeleton—skeleton-based representations have emerged as a compact and interpretable form, enabling efficient motion modeling and strong generalization across subjects and environments.

The introduction of spatio-temporal graph convolutional networks (ST-GCNs)~\cite{yan2018stgcn} 
and their successors~\cite{chen2021ctrgcn, chi2022infogcn, lee2023hierarchically, shi2019twoagcn, wang20233mformer, zhou2024blockgcn} has significantly advanced skeleton-based HAR by jointly modeling spatial and temporal dependencies among body joints. 
Despite these advances, most methods remain \textit{body-centric}, emphasizing large-scale body motions while overlooking fine hand articulations that are essential to fine-grained recognition. 
Although unified graph models such as SkeleT~\cite{yang2024expressive} integrate body and hand joints within a holistic topology, 
dominant body dynamics often overshadow subtle hand cues due to inherent scale and feature imbalance, limiting specialization and robustness. 

In practice, hand skeletons—comprising small-scale joints and fine-grained articulations—are highly susceptible to occlusion and estimation noise, 
leading to a fundamental \textit{reliability asymmetry} between body and hand modalities: stable body joints versus noisy or missing hand joints. 
Robust recognition thus calls for a formulation that explicitly models this asymmetry and adaptively regulates modality interactions.
We address this by introducing three components and unifying them into a reliability-aware dual-stream body–hand framework, which we term \textbf{BHaRNet} (\textbf{B}ody–\textbf{H}and \textbf{a}ction \textbf{R}ecognition \textbf{Net}work).

\begin{figure}
    \centering
    \begin{minipage}[t]{0.48\columnwidth}
        \centering
        \includegraphics[width=\linewidth]{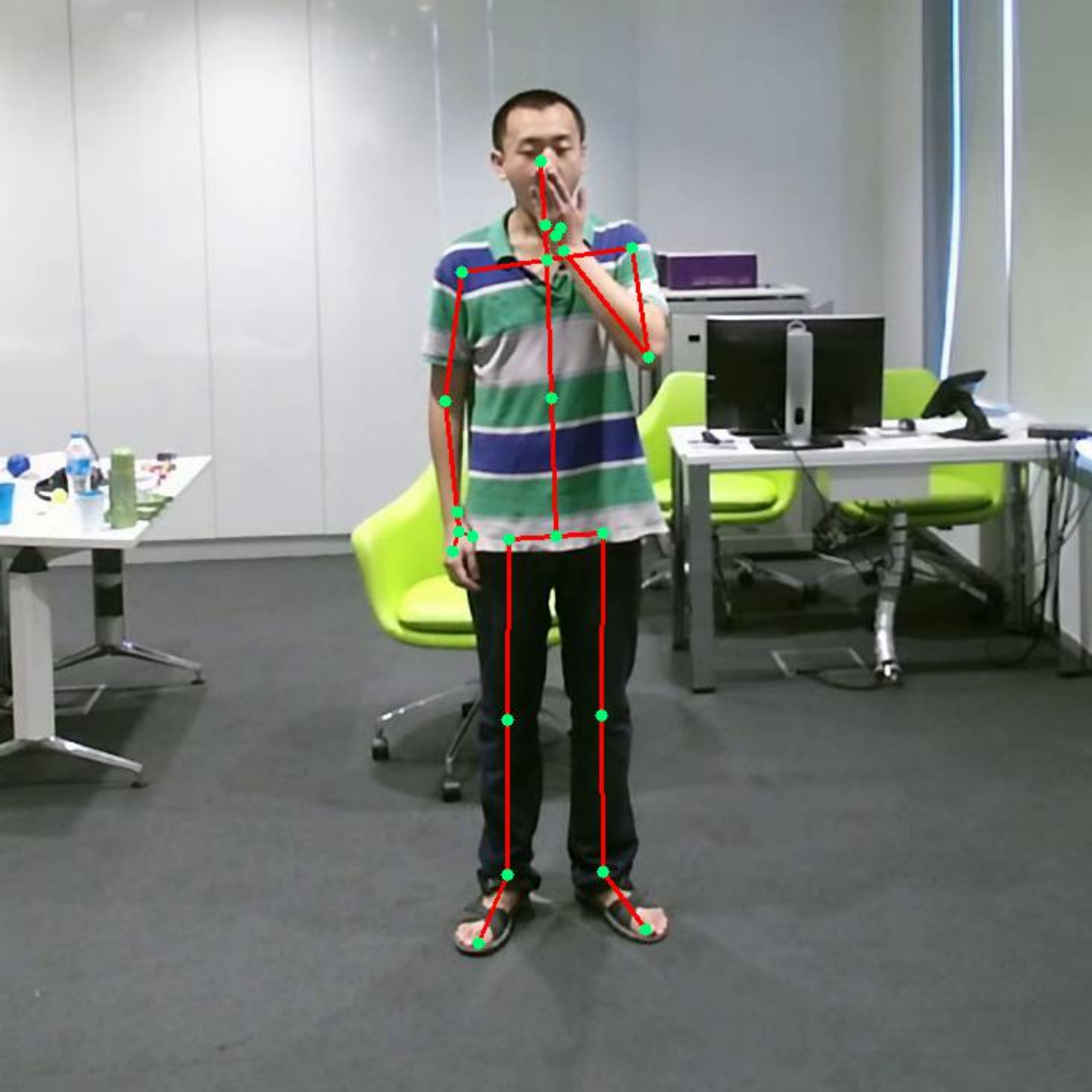}
    \end{minipage}
    \hfill
    \begin{minipage}[t]{0.48\columnwidth}
        \centering
        \includegraphics[width=\linewidth]{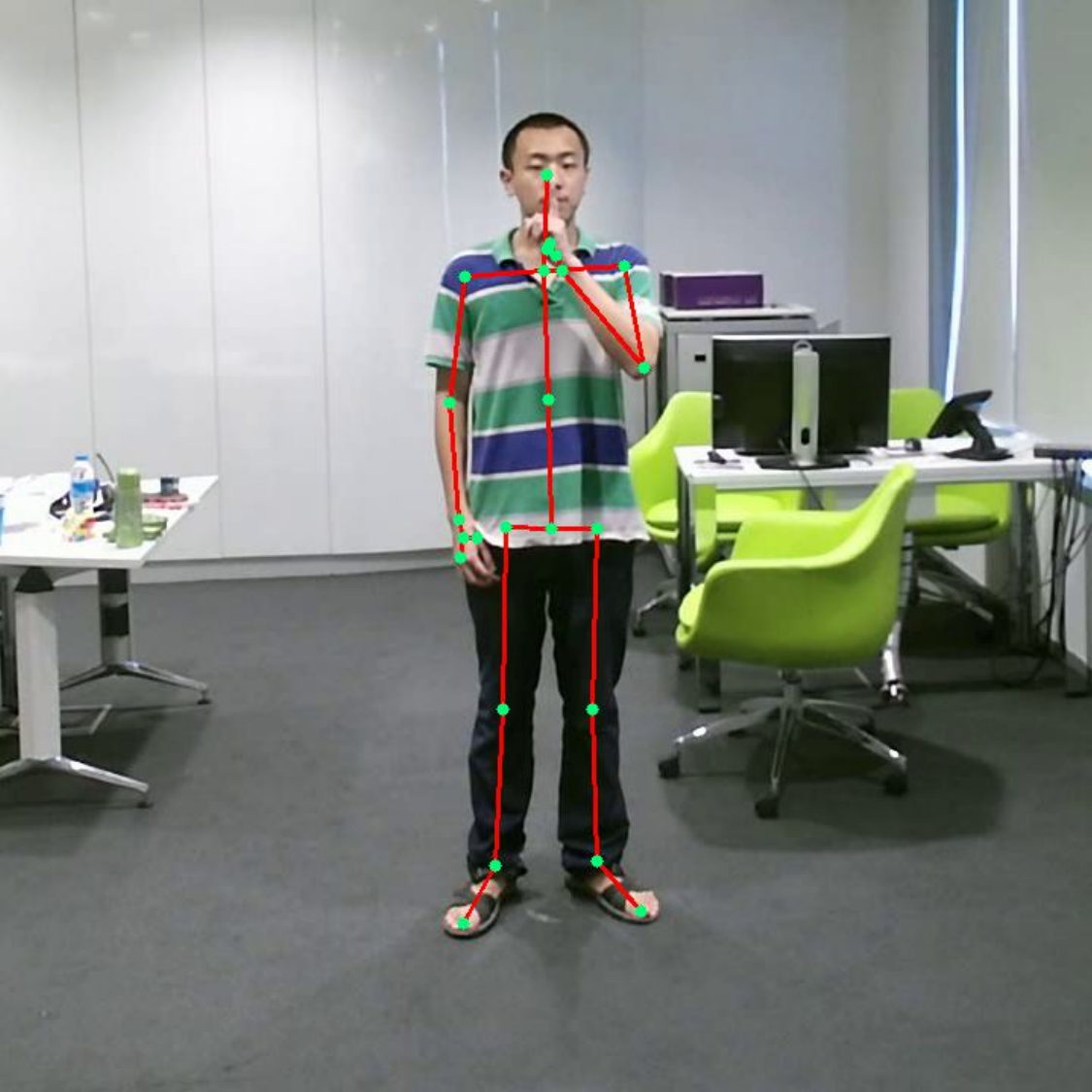}
    \end{minipage}
    \caption{Visualization of two hand-centric actions—``Yawn”(left) and ``Hush”(right)—cropped frames from NTU RGB+D with body skeleton.
Both representations share nearly identical body postures, indicating strong global pose similarity across distinct hand gestures. This highlights the challenge of distinguishing fine-grained actions using only body skeletons and motivates the need for reliability-aware hand modeling.}
    \label{fig:yawn_hush}
\end{figure}
\noindent\textbf{Calibration-free Learning.}
Earlier hand-pose frameworks often applied canonical-space alignment to normalize hand coordinates, assuming accurate joint location~\cite{lee2025toward, li2019spatial, shamil2024utility}.
Such transformations often amplify noise under occlusion or motion blur, propagating local errors throughout the skeleton. 
To address this, we adopt a calibration-free design by eliminating canonical-space transformations and allowing the model to operate directly in the native coordinate space. This design encourages learning unified skeleton representations while mitigating error propagation.

\noindent\textbf{Probabilistic Noisy-OR Fusion.}
Deterministic fusion schemes typically assume equal reliability across modalities, leading to unstable aggregation when hand cues are uncertain or missing. 
We introduce a Noisy-OR fusion that models reliability asymmetry between modalities, functioning as a gate that amplifies strong, consistent evidence while suppressing unreliable signals, all without relying on explicit confidence supervision. 
This probabilistic mechanism stabilizes reliability-aware aggregation across body and hand streams and improves robustness, particularly for fine-grained, hand-driven actions under noisy or missing keypoints.

\noindent\textbf{Multi-modal Ensemble Beyond Skeletons.}
The conventional intra-skeleton multi-modal ensemble—integrating four skeleton modalities (Joint, Bone, Joint Motion, and Bone Motion)—has proven effective for modeling spatial and temporal cues within skeleton-based recognition. 
We further extend this principle beyond the skeleton domain: inspired by MMNet~\cite{bruce2022mmnet}, 
a body-guided modulation strategy is employed in which joint-motion features dynamically condition RGB feature learning, 
bridging structural and visual representations in a unified cross-modal formulation.

This article builds on our conference work~\cite{cho2025body} and extends it into a unified probabilistic dual-stream framework. 
Section~\ref{sec:background} revisits the deterministic body–hand baselines and preprocessing pipeline, 
while Section~\ref{sec:method} presents the generalized probabilistic framework. 
Section~\ref{sec:experiments} reports comprehensive evaluations, including new hand-centric benchmarks, ablations, and robustness and efficiency analyses. 
Implementation details and full quantitative results are deferred to the Appendix.

\noindent\textbf{Contributions.}
\begin{itemize}
  \item Generalized probabilistic dual-stream framework: integrates calibration-free skeleton learning, reliability-aware fusion, and cross-modal ensemble under a unified formulation that models asymmetric reliability between body and hand modalities.
  \item Noisy-OR fusion: introduces a probabilistic mechanism that stabilizes reliability-aware fusion without explicit confidence supervision.
  \item Intra- to cross-modal ensemble: extends intra-skeleton motion cue integration (Joint, Bone, Joint Motion, Bone Motion) to the visual modality, enabling unified skeleton–RGB learning in a cross-modal formulation.
  \item Comprehensive validation and generalization analysis: includes a new hand-centric benchmark (NTU-Hand 11/27), extended ablations on calibration-free and Noisy-OR components, noise robustness under frame-drop conditions, and cross-modal evaluations across multiple benchmarks, demonstrating consistent improvements under diverse conditions.
\end{itemize}
\section{Related Work}
\label{sec:related}
\subsection{Skeleton-based Action Recognition}
Skeleton-based action recognition has progressed through several architectural paradigms, from sequence models to graph- and attention-based networks. 
Early approaches relied on RNNs~\cite{du2015hierarchical} to capture temporal dynamics, but they were limited in explicitly modeling spatial relations among joints. 
Subsequent works explored convolutional architectures by encoding joint trajectories into image-like representations and applying CNNs~\cite{duan2022poseconv, hachiuma2023unified}, which improved efficiency but still treated the underlying skeletal structure only implicitly.

The introduction of spatio-temporal graph convolutional network (ST-GCN)~\cite{yan2018stgcn} marked a key shift by representing joints as graph nodes and kinematic connections as edges, enabling joint modeling of spatial and temporal dependencies. 
Building on this formulation, a series of GCN-based methods~\cite{chen2021ctrgcn, chi2022infogcn, lee2023hierarchically, liu2025protogcn, shi2019twoagcn, zhou2024blockgcn} refined graph topology design, aggregation operators, and multi-stream representations to enhance robustness and discriminative power. 
Furthermore, graph-based Transformer frameworks~\cite{abdelfattah2024maskclr, duan2023skeletr, pang2022igformer, wang20233mformer} have incorporated self-attention to capture long-range dependencies and global motion patterns.

While most of these models remain predominantly body-centric, recent work such as SkeleT~\cite{yang2024expressive} integrates body, hand, and foot keypoints into a unified graph to capture holistic human dynamics. 
However, in such unified formulations, global body motions may still overshadow subtle hand articulations, motivating the need for frameworks that explicitly account for modality-specific reliability and specialization, as pursued in this work.

\subsection{Multi-modal Action Recognition}
Early graph-based approaches~\cite{liu2020disentangling, shi2019twoagcn} primarily relied on two input modalities—joint and bone representations. 
Later works~\cite{chen2021ctrgcn, cheng2020shift, yang2024expressive} expanded this paradigm to four modalities by incorporating temporal dynamics through joint motion (JM) and bone motion (BM). 
Further extensions~\cite{chi2022infogcn, liu2025protogcn} explored six-stream variants that jointly capture multiple spatial and temporal cues. 
More recently, 3MFormer~\cite{wang20233mformer} demonstrated that hypergraph-based formulations can flexibly aggregate multi-scale spatial and temporal relations, 
validating the importance of diverse modality fusion within skeleton-based recognition. 

Beyond purely skeletal data, a growing body of research integrates visual modalities to exploit appearance cues and context. 
Recent approaches~\cite{abdelkawy2025epam, bruce2022mmnet, duan2022poseconv, reilly2024justvit} couple pose and RGB representations through architectural alignment or feature fusion. 
Among these, MMNet~\cite{bruce2022mmnet} proposed a body-guided modulation strategy, 
using body features to weight RGB activations, effectively transferring structural priors into visual representations. 
These advances motivate our probabilistic dual-stream ensemble that unifies skeleton modalities and visual cues under a reliability-aware formulation.

\subsection{Body–Hand Coordination in Related Fields}
Modeling body–hand dependencies has been actively investigated in neighboring domains such as sign-language recognition, motion forecasting, and egocentric understanding.  
For instance, UNI-SIGN~\cite{li2025unisign} embeds full-body, hand, and facial keypoints into a shared latent space for sign recognition, 
while ExpForecastAI~\cite{ding2024expforcasteai} and REWIND~\cite{lee2025rewind} focus on future pose alignment and temporal coherence prediction.  
Distinct from these methods, we preserve body and hand as separate experts and employ lightweight cross‑attention, avoiding the need for the shared embedding or the high-precision localization.

\subsection{Canonical-space Normalization and Viewpoint Robustness}
\label{subsec:related_canonical}
Canonical-space normalization has been widely adopted in gesture and hand-pose estimation~\cite{lee2025toward, li2019spatial, shamil2024utility} to align local hand coordinates and reduce viewpoint variation, often by transforming hand joints into an egocentric or canonical frame. 
Such transformations have been shown to improve recognition in controlled, close-range settings~\cite{li2019spatial, shamil2024utility}, but they presuppose precise hand joint estimation and stable local geometry—an assumption rarely satisfied in large-scale, third-person HAR datasets where hand joints are small, frequently occluded, or missing. 
Under these conditions, canonical mapping can amplify estimation noise and propagate local errors across the skeleton, ultimately degrading recognition performance.

In body action recognition, early skeleton-based methods relied on canonical mapping to handle viewpoint variability~\cite{liu2020disentangling, yan2018stgcn}, whereas more recent architectures favor data-driven robustness via sequence-level random rotations of 3D skeletons during training~\cite{chen2021ctrgcn, myung2024degcn, zhang2020semantics}. 
These works demonstrate that strong performance and view invariance can be achieved without a fixed canonical space. 
Our work follows this latter direction in the more challenging body--hand setting, where hand estimates are substantially noisier than body joints: instead of canonical alignment, we adopt a calibration-free design in the native coordinate space and explicitly model reliability asymmetry between body and hand modalities, as detailed in Section~\ref{sec:method}.

\subsection{Probabilistic Fusion and Reliability-aware Modeling}
The Noisy-OR model~\cite{pearl1988probabilistic} has long served as a fundamental operator in probabilistic reasoning, 
modeling the likelihood of an event triggered by multiple independent causes.  
Its modern adaptations in deep learning~\cite{tian2019uno, wu2015deep} 
enable reliability-aware multi-instance aggregation and uncertainty modeling in perception and fusion tasks.  
Unlike deterministic averaging or concatenation, Noisy-OR aggregation naturally encodes asymmetric confidence between sources, suppressing unreliable signals while retaining strong evidence.  
In this work, we instantiate such probabilistic reasoning in a dual-stream setting by adopting a Noisy-OR fusion mechanism to govern reliability-aware aggregation across body-hand modalities.

\section{Background and Foundational Frameworks}
\label{sec:background}
This section formalizes the notations used throughout the paper, describes the preprocessing pipeline for body–hand skeleton construction, and revisits the deterministic dual-stream architectures that form the basis of our probabilistic framework.

\subsection{Notation and Problem Definition}
We denote a body–hand skeleton sequence as
\begin{align}
    X_B \in \mathbb{R}^{C \times T \times V_B}, \qquad
X_H \in \mathbb{R}^{C \times T \times V_H},
\end{align}
where $C$ denotes coordinate channels (x, y, z), $T$ the temporal length, and $V_B, V_H$ the number of joints for body and hand, respectively.
Given $K$ action classes, the goal is to learn
\begin{align}
    f_\theta:(X_B, X_H)\rightarrow y,\qquad y \in \mathbb{R}^K,
\end{align}
robustly under noisy or missing keypoints.

A characteristic challenge of body–hand modeling is the reliability asymmetry
between modalities: body joints are generally stable, whereas hand joints suffer from occlusion, blur, and frequent estimation failures. This motivates the reliability-aware learning introduced in Section~\ref{sec:method}.

\subsection{Preprocessing Pipeline Overview}
We adopt a preprocessing pipeline used in our baseline framework:
hand keypoint extraction (Mediapipe~\cite{lugaresi2019mediapipe}), temporal smoothing, zero-masked dummy nodes for topology consistency, canonical transformation, hip-centered normalization, and temporal resampling. 
These steps form a deterministic preprocessing pipeline identical to our conference baseline~\cite{cho2025body},
except that in this work we remove the canonical transform while preserving the rest (Section~\ref{sec:method:canonical}).
Additional preprocessing details are provided in Appendix~\ref{appendix:preprocess}.

\subsection{Foundational Dual-Stream Architectures}
\label{sec:background:baseline}
Our conference framework~\cite{cho2025body} introduced a deterministic dual-stream design to jointly model body dynamics and hand articulations. 
We briefly summarize its key components as the foundation for the generalized probabilistic framework proposed in this work.

\subsubsection{Dual-Stream Training with Complementary Loss}
The baseline dual-stream network consists of two spatio-temporal GCN backbones, 
each specialized for either the body or the hand modality. 
Given their logit outputs $\hat{y}_B$ and $\hat{y}_H$, 
the training objective combines an individual loss $\mathcal{L}_{\text{idv}}$ and a complementary loss $\mathcal{L}_{\text{cpl}}$:
\begin{gather}
    \mathcal{L}_{\text{idv}} = \mathrm{CE}(\hat{y}_B, l) + \mathrm{CE}(\hat{y}_H, l),\\
\mathcal{L}_{\text{cpl}} = \mathrm{CE}\!\left(\tfrac{1}{2}(\hat{y}_B+\hat{y}_H),\, l\right),\\
\mathcal{L}_{\text{total}} = \lambda_{\text{idv}}\mathcal{L}_{\text{idv}} + \lambda_{\text{cpl}}\mathcal{L}_{\text{cpl}}.
\end{gather}
Here, CE is the standard cross-entropy with softmax.

This joint training scheme encourages inter-modality collaboration while preserving modality-specific specialization.
To alleviate domain discrepancy between estimated hand skeletons and body joints, a canonical-space transformation was applied to normalize hand coordinates. 
While this preprocessing stabilized training in some cases, it also amplified noise propagation 
when hand joint estimations were inaccurate, motivating the calibration-free design in Section~\ref{sec:method:canonical}.

\subsubsection{Cross-Attention Mechanisms}
\begin{figure*}[t]
    \centering
    \includegraphics[width=0.95\textwidth]{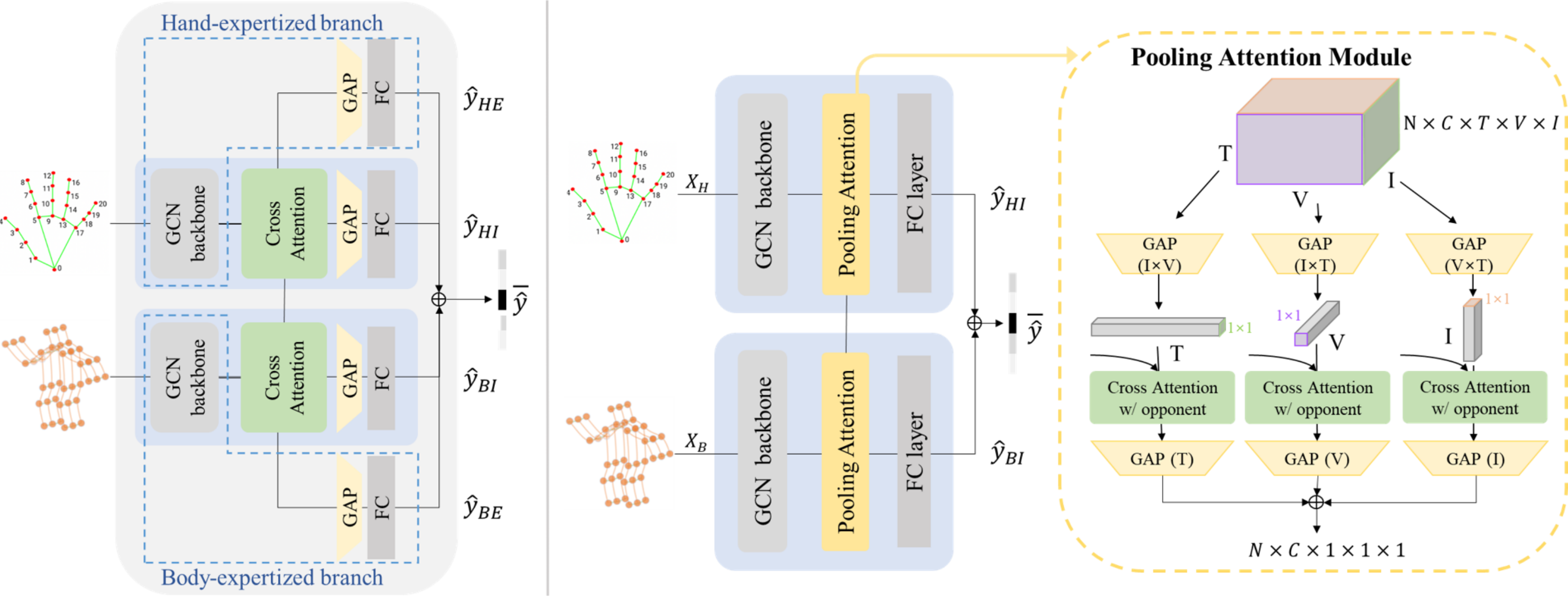}
    \caption{Overview of the dual-stream architectures.
    \textbf{Left:} BHaRNet-P with interactive body (BI) and hand (HI) branches connected via lightweight cross-attention.
    \textbf{Right:} BHaRNet-E with additional expertized branches (BE, HE) that preserve modality-specific cues while sharing context through the interactive branches.
    The corresponding branch–loss configurations for deterministic baseline and probabilistic framework are summarized in Table~\ref{tab:branch-loss-map}.
    }
    \label{fig:overview}
\end{figure*}

To enhance information exchange between body and hand streams, 
a cross-attention mechanism was incorporated at the feature level. 
The \textbf{BHaRNet-P} variant employed two interactive branches—a body-interactive branch (BI) and a hand-interactive branch (HI)—which exchange features via a pooling attention module (Fig.~\ref{fig:overview}).
The \textbf{BHaRNet-E} variant further introduced expertized branches—a body-expert branch (BE) and a hand-expert branch (HE)—to balance modality-specific specialization and cross-modality communication.  
During training, both interactive and expertized branches participated in loss computation, 
while at inference, a lightweight configuration using only the expertized branches achieved high efficiency.

\begin{table*}
    \centering
    \resizebox{0.99\linewidth}{!}{
    \begin{tabular}{lccccc}
        \toprule
        Model & Variants & Branches & $\mathcal{L}_{\text{idv}}$ & $\mathcal{L}_{\text{cpl}}$ & $\mathcal{L}_{\text{nor}}$ \\
        \midrule
        \multirow{2}{*}{Deterministic Baseline} & P & BI, HI & CE($y_{BI}$)+CE($y_{HI}$) & CE(avg(BI,HI)) & -- \\
        & E & BI,HI,BE,HE & CE($y_{BI}$)+CE($y_{HI}$) & CE(avg(all)) & -- \\
        \midrule
        \multirow{2}{*}{Probabilistic Framework} & P & BI, HI & CE($y_{BI}$)+CE($y_{HI}$) & CE(avg(BI,HI)) & CE(Noisy-OR($\sigma(y_{BI}),\sigma(y_{HI})$)) \\
        & E & BI,HI,BE,HE & CE($y_{BI}$)+CE($y_{HI}$) & CE(avg(all)) & CE(Noisy-OR($\sigma(y_{BE}),\sigma(y_{HE})$)) \\
        \bottomrule
    \end{tabular}}
    \caption{Branch–loss–output mapping for deterministic baseline and probabilistic framework. 
    BI/HI: body/hand interactive branches; BE/HE: body/hand expertized branches.  
    $\sigma$: sigmoid activation.  
    Noisy-OR is applied element-wise.}
    \label{tab:branch-loss-map}
\end{table*}

\subsubsection{Deterministic Skeleton--RGB Ensemble}
Beyond skeleton-only recognition, the deterministic framework was extended to incorporate RGB cues for appearance-based reasoning.  
Following the principle of MMNet~\cite{bruce2022mmnet}, 
the body-expert feature from the joint modality, $f_{BE}$, 
served as a spatial weighting signal for RGB activations:
\begin{align}
    f'_{\text{RGB}} = f_{\text{RGB}} \odot \mathrm{weight}(f_{BE}),
\end{align}

where $\mathrm{weight}(\cdot)$ denotes a learnable transformation producing region-wise importance maps.
This modulation transferred structural priors from skeleton features to RGB representations, 
aligning spatial saliency with articulated motion. 
Let $J$ and $B$ denote the joint and bone modalities, respectively, the final predictions were obtained by deterministic weighted summation:
\begin{align}
    \hat{y}_{\text{final}} = w_J \hat{y}_J + w_B \hat{y}_B + w_{\text{RGB}}\hat{y}_{\text{RGB}}. 
\end{align}

This ensemble provided a strong baseline for integrating structural and appearance cues under a unified deterministic formulation, with ensemble weights fixed as  $(w_J:w_B:w_{\text{RGB}})$ = (1:1:1).

The above architectures form the deterministic foundation upon which our generalized probabilistic dual-stream framework is built. 
In the next section, we reformulate these deterministic components into a reliability-aware probabilistic learning framework 
that explicitly models uncertainty and extends multi-modal fusion into the temporal domain.

\section{Generalized Probabilistic Dual-Stream Framework}
\label{sec:method}
We now present a generalized probabilistic dual-stream framework that extends the deterministic baseline (Section~\ref{sec:background}) by introducing reliability-aware learning and structured multi-modal integration. 
The framework preserves the efficiency and deterministic inference of the original design while embedding probabilistic modeling into the training objective to improve robustness and specialization under uncertain skeleton data. Fig.~\ref{fig:overview} illustrates the dual-stream architectures (BHaRNet-P/E) that serve as the backbone for both the deterministic baseline and our probabilistic extension.

\subsection{Overview}
The framework consists of three key components: 
(1) representation learning without canonical transformation, 
(2) reliability-aware probabilistic learning with Noisy-OR loss, and 
(3) multi-modal ensemble incorporating temporal cues.  
The total training objective combines deterministic and probabilistic supervision as:
\begin{equation}
    \mathcal{L}_{\text{total}} = 
    \lambda_{\text{idv}}\mathcal{L}_{\text{idv}} + 
    \lambda_{\text{cpl}}\mathcal{L}_{\text{cpl}} + 
    \lambda_{\text{nor}}\mathcal{L}_{\text{nor}},
\end{equation}
where $\mathcal{L}_{\text{idv}}$ and $\mathcal{L}_{\text{cpl}}$ represent individual and complementary losses from the baseline, and $\mathcal{L}_{\text{nor}}$ denotes the proposed probabilistic Noisy-OR regularization term.  
During inference, predictions remain deterministic, obtained by logit summation across branches, ensuring efficiency identical to the baseline.

\subsection{Representation Learning without Canonical Transformation}
\label{sec:method:canonical}
\begin{figure}
    \centering
    \begin{minipage}[t]{0.47\linewidth}
        \centering
        \includegraphics[width=\linewidth]{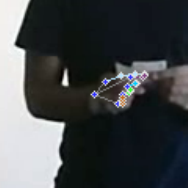}%
        \\[-2pt]
        \small (a) RGB crop with estimated hand (accurate case).\\[4pt]
        \includegraphics[width=\linewidth]{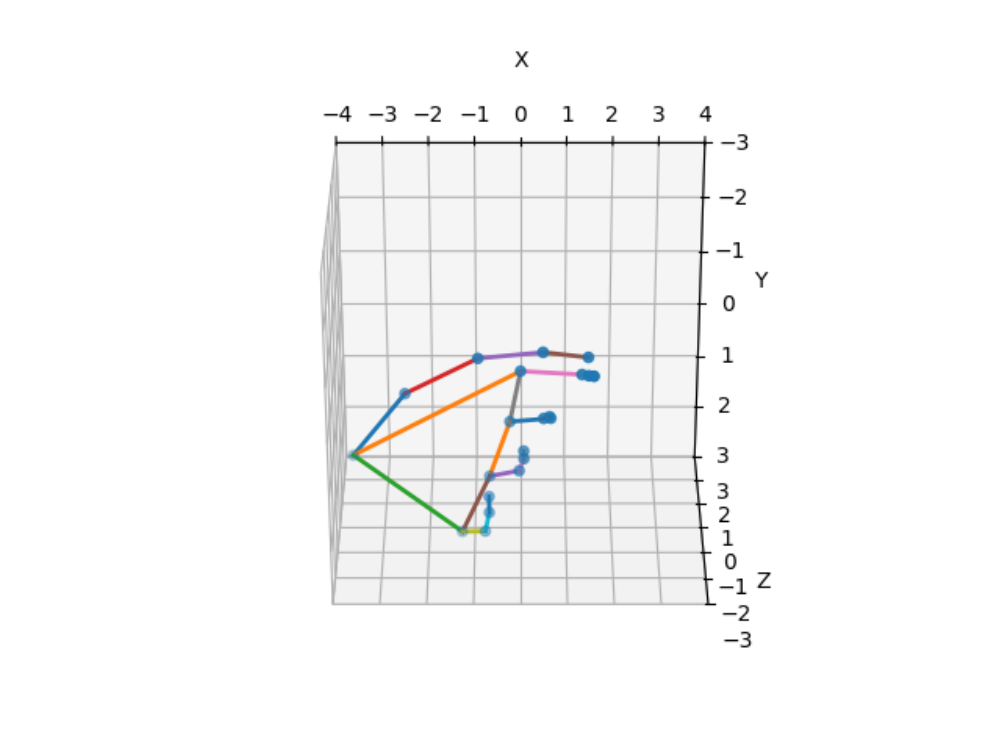}%
        \\[-2pt]
        \small (b) 3D hand skeleton in native coordinates.\\[4pt]
        \includegraphics[width=\linewidth]{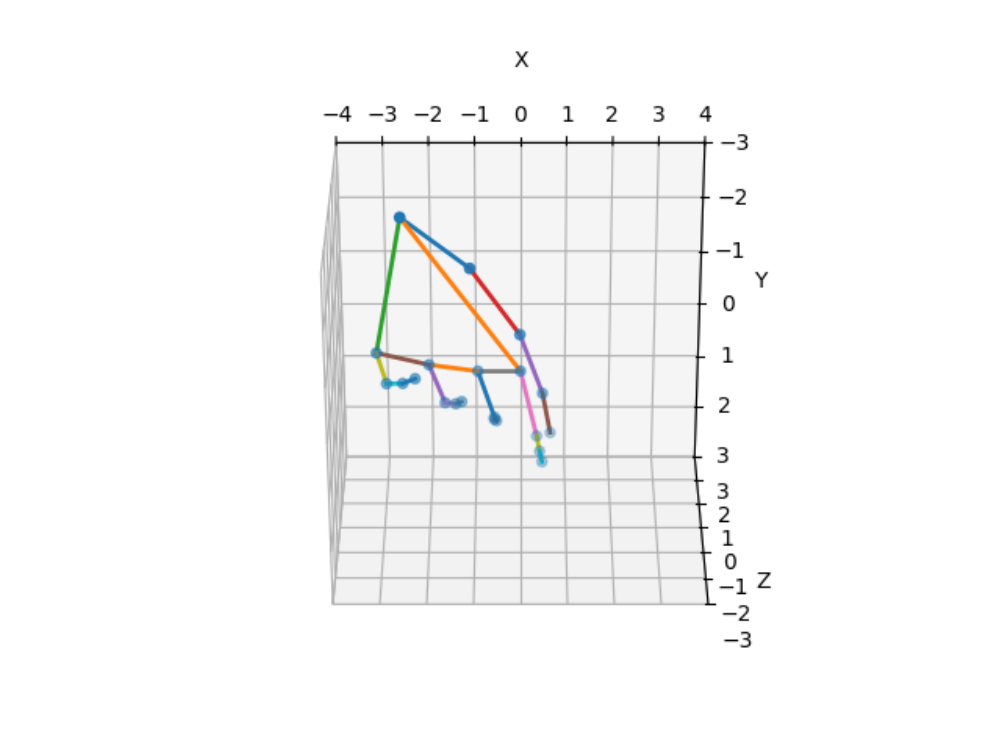}%
        \\[-2pt]
        \small (c) 3D hand skeleton after canonical-space transformation.
    \end{minipage}
    \hfill
    \begin{minipage}[t]{0.47\linewidth}
        \centering
        \includegraphics[width=\linewidth]{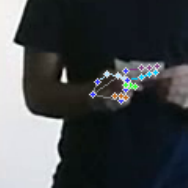}%
        \\[-2pt]
        \small (d) RGB crop with estimated \\
        hand (noisy case).\\[4pt]
        \includegraphics[width=\linewidth]{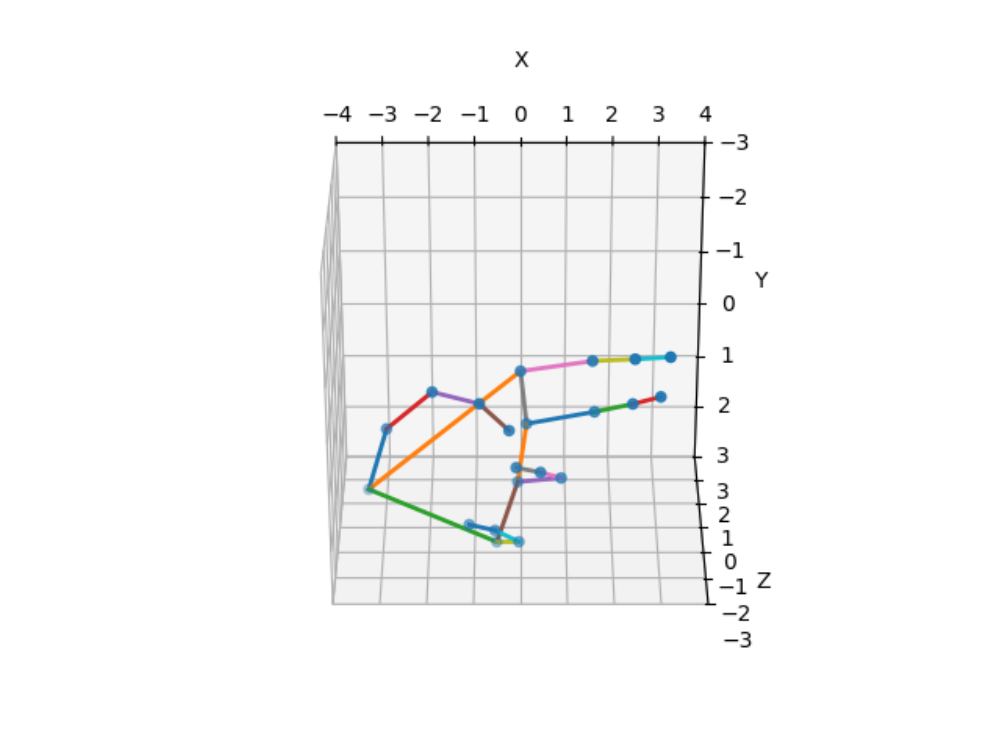}%
        \\[-2pt]
        \small (e) 3D hand skeleton in native coordinates.\\[4pt]
        \includegraphics[width=\linewidth]{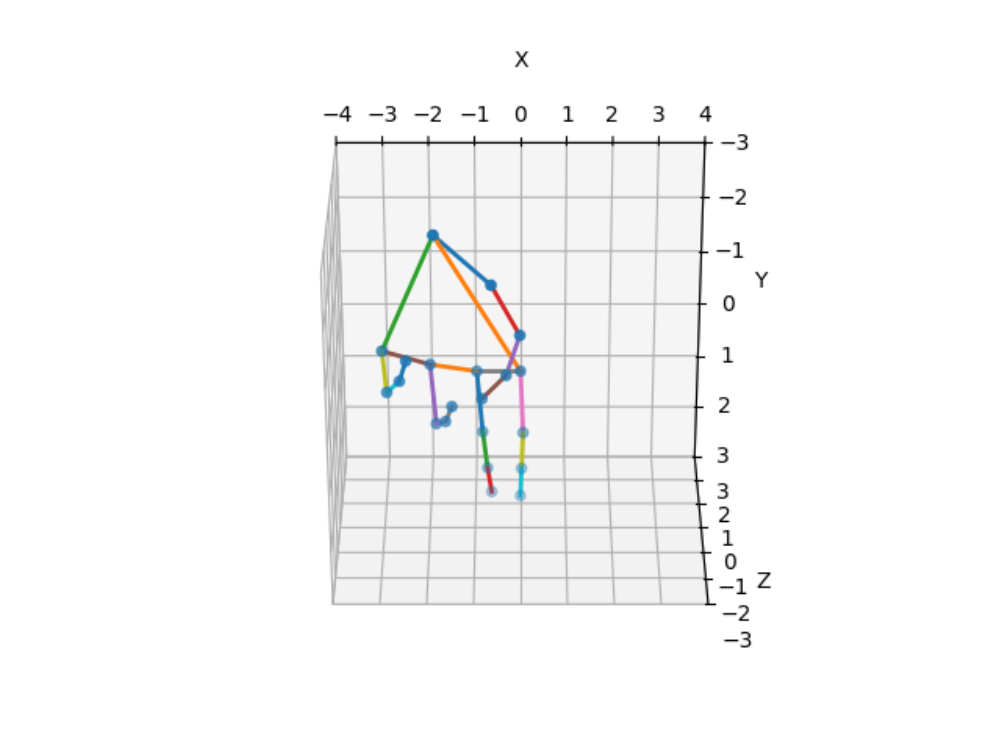}%
        \\[-2pt]
        \small (f) 3D hand skeleton after canonical-space transformation with amplified distortion.
    \end{minipage}

    \caption{
    Motivating example for the calibration-free representation learning used in our generalized framework.
    We visualize two consecutive frames at 30 fps from a single sequence.
    Left: frame 17 with a accurate hand estimate.
    Right: frame 18 where the index and middle fingers are corrupted by noise.
    Rows show (top) RGB crops with hand estimation,
    (middle) native 3D hand skeletons, and
    (bottom) 3D hand skeletons after canonical-space transformation.
    All 3D views share the same viewpoint and axis scales for fair comparison.
    In the native space, the overall hand configuration remains stable
    except around the noisy index and middle joints.
    After canonical-space transformation,
    local noise propagates to the entire hand,
    causing large joint-wise displacements and noticeable shape distortion.
    }
    \label{fig:noise_canonical}
\end{figure}

Previous body–hand frameworks often relied on canonical-space normalization to align hand coordinates across subjects.
While such calibration reduces inter-subject variance, it assumes accurate keypoint estimation—a condition frequently violated in large-scale third-person HAR data where hand joints suffer from occlusion and motion blur.  
Errors in local hand joints thus propagate globally through the transformation, amplifying noise (Fig.~\ref{fig:noise_canonical}).

We eliminate this canonical step and directly learn from native skeleton coordinates.
This design preserves local spatial consistency, prevents error propagation from unreliable joints, 
and keeps body and hand in the same coordinate system, allowing the model to implicitly learn cross-scale correspondences.
In practice, we found that the calibration-free setting yields more stable training dynamics under hand occlusion and motion blur, as analyzed in Section~\ref{sec:experiments}.

\begin{figure*}
    \centering
    \includegraphics[width=0.58\textwidth]{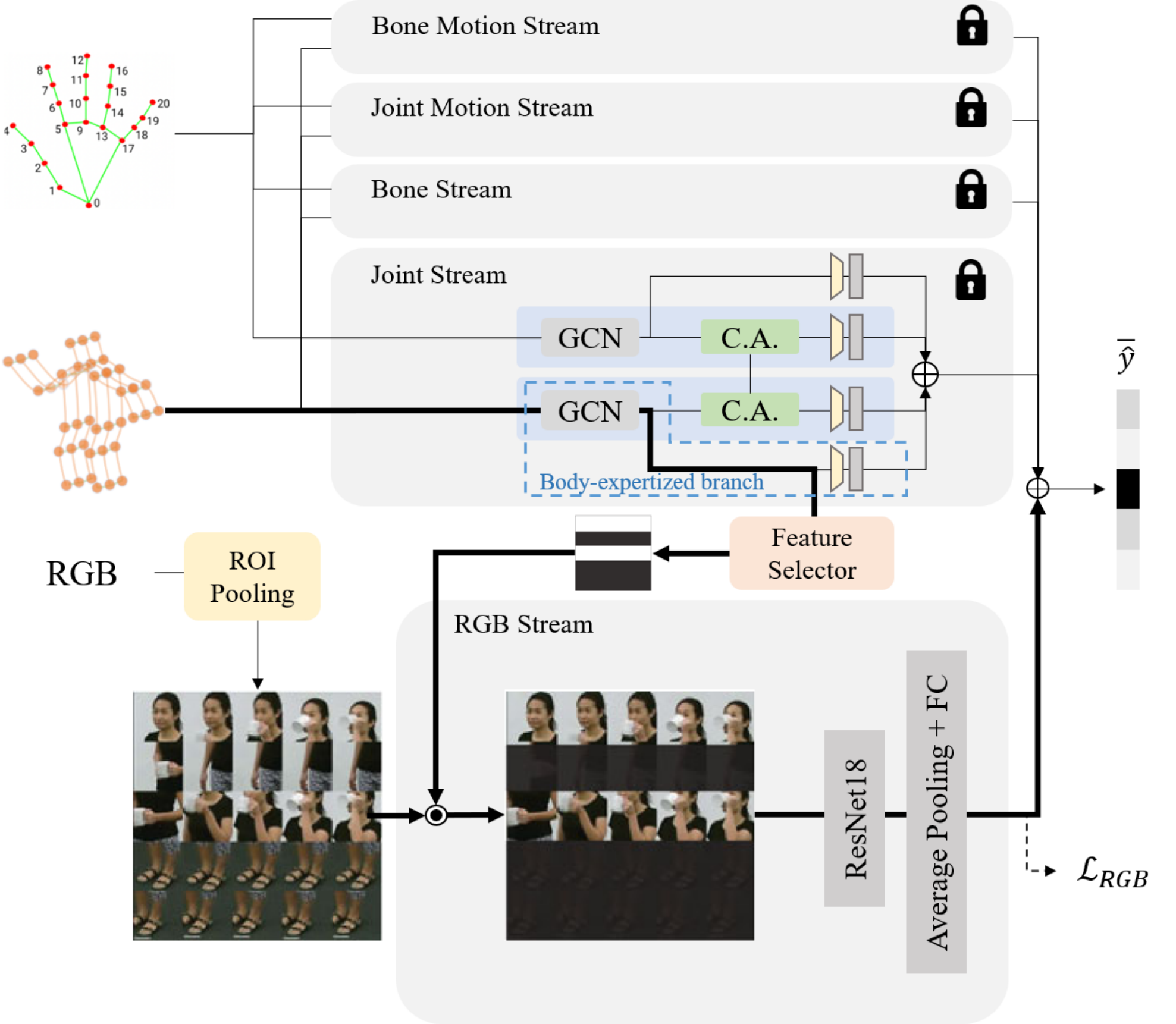}
    \caption{Schematic of our BHaRNet-M. We integrate BHaRNet-E for skeletal streams and add an RGB stream with its own training path (bold lines). The RGB branch receives body-joint guidance from the body-expertized branch, focusing the visual feature extractor on relevant spatio-temporal regions.}
    \label{fig:bharnet_m}
\end{figure*}
\subsection{Reliability-Aware Probabilistic Learning with Noisy-OR Loss}
The deterministic loss formulation in Section~\ref{sec:background} implicitly assumes equal reliability across all branches.
In practice, hand-related branches are often unstable due to missing or inaccurate keypoints, causing inconsistent learning signals.
To address this, we introduce a reliability-aware term based on the Noisy-OR operator, 
which aggregates branch-wise evidence in a way that favors configurations where at least one branch confidently supports the ground-truth class.

\paragraph{Probabilistic formulation.}
Given branch-level logits $y_i \in \mathbb{R}^K$, we first compute
per-class scores via a sigmoid:
\begin{equation}
    p_{i,k} = \sigma(y_{i,k}), \quad k = 1, \dots, K,
\end{equation}
which maps each logit to $[0,1]$ and stabilizes inter-branch scale differences.
We then aggregate these scores across branches using an element-wise Noisy-OR operator:
\begin{equation}
    p_{\text{nor},k} = 1 - \prod_i \bigl(1 - p_{i,k}\bigr).
\end{equation}
Here, $p_{\text{nor},k}$ becomes large if at least one branch assigns high confidence to class $k$, 
while unreliable branches with low scores contribute little. 

We do not interpret $p_{\text{nor}} = (p_{\text{nor},1}, \dots, p_{\text{nor},K})$ as a normalized probability distribution, but as per-class evidence scores. 
We then feed $p_{\text{nor}}$ into the softmax-based cross-entropy:
\[
    \mathcal{L}_{\text{nor}} = \mathrm{CE}(p_{\text{nor}}, l).
\]
In other words, Noisy-OR provides a differentiable pooling of branch-wise evidence before computing a standard single-label cross-entropy loss.
We use $\mathcal{L}_{\text{nor}}$ as a training-time regularizer that encourages at least one branch to confidently support the correct class, 
without changing the deterministic inference rule.

\paragraph{Variant-specific application.}
Table~\ref{tab:branch-loss-map} summarizes the branch–loss mapping for both deterministic and probabilistic variants.
In BHaRNet-P, the Noisy-OR term is defined over the interactive branches BI and HI:
\begin{equation}
    p_{\text{nor}}^{(P)} = 1 - \bigl(1 - \sigma(y_{BI})\bigr) \odot \bigl(1 - \sigma(y_{HI})\bigr),
\end{equation}
and $\mathcal{L}_{\text{nor}}$ is computed from the softmax-normalized Noisy-OR scores.
This encourages the model to rely on whichever interactive branch remains reliable, while mitigating the effect of a corrupted counterpart.

In BHaRNet-E, expert branches BE and HE are designed to specialize in modality-specific reasoning, 
while interactive branches BI and HI maintain cross-modal context through the complementary loss.
Here, the Noisy-OR term is defined only over the expert branches:
\begin{equation}
    p_{\text{nor}}^{(E)} = 1 - \bigl(1 - \sigma(y_{BE})\bigr) \odot \bigl(1 - \sigma(y_{HE})\bigr),
\end{equation}
so that either body or hand expert can dominate the decision when confident.
This design matches the lightweight inference mode, which uses only BE and HE.

\paragraph{Interpretation and independence.}
Strict statistical independence between branches is not guaranteed in our setting,
since the branches share upstream encoders and supervision, and BI/HI exchange information via cross-attention.
However, in the expertized configuration (BHaRNet-E), BE/HE do not directly attend to each other and are encouraged to focus on modality-specific cues.
We therefore interpret the Noisy-OR operator as an approximate pooling of partially independent experts rather than a fully generative probabilistic model, 
and use $\mathcal{L}_{\text{nor}}$ as a training-time regularizer that biases learning toward configurations where at least one expert confidently supports the correct class.

\subsection{Multi-Modal Ensemble with Temporal Cues}
We further generalize the deterministic ensemble (Section~\ref{sec:background}) by incorporating temporal motion modalities under the same logit-sum rule.
The ensemble remains deterministic and parameter-free, extending spatial skeleton cues (Joint, Bone) with temporal dynamics (Joint Motion, Bone Motion), and integrating RGB appearance cues in a unified logit-space formulation.

\begin{table*}
    \centering
    \caption{Accuracy (\%), FLOPs, and Parameter Comparison with State-of-the-Art Methods on NTU RGB+D 60/120 and N-UCLA benchmarks in Skeleton-based action recognition. “-” indicates the experimental results are not provided in the reference, and “*” for the result based on using public codes. Best and second-best results are highlighted in \textbf{bold} and \underline{underline}, respectively. BHaRNet-*$\dagger$ denote our conference baselines, while “Ours” rows correspond to the proposed probabilistic variants.}
    \label{tab:exp_skel}
    \resizebox{0.8\linewidth}{!}{
    \begin{tabular}{lccccccc}
    \toprule
    \multirow{2}{*}{Method} & 
    \multicolumn{2}{c}{NTU 60} &
    \multicolumn{2}{c}{NTU 120} &
    {N-UCLA} &
    \multirow{2}{*}{GFLOPs} & 
    \multirow{2}{*}{Params(M)} \\ \cmidrule(lr){2-3}\cmidrule(lr){4-5}\cmidrule(lr){6-6}
    
     &  X-Sub & X-View & X-Sub & X-Set & X-View & & \\
    \midrule
    CTR-GCN~\cite{chen2021ctrgcn} & 92.4 & 96.8 & 88.9 & 90.6 & 96.5 & 7.9 & 5.8 \\
    InfoGCN~\cite{chi2022infogcn} & 93.0 & 97.1 & 89.8 & 91.2 & 97.0 & 10.0* & 9.4\\
    PoseConv3D~\cite{duan2022poseconv} & 94.1 & 97.1 & 86.9 & 90.3 & - & 31.8 & \underline{4.0} \\
    BlockGCN~\cite{zhou2024blockgcn} & 93.1 & 97.0 & 90.3 & 91.5 & 96.9 & 6.5 & 5.2\\
    DeGCN~\cite{myung2024degcn} & 93.6 & 97.4 & 91.0 & 92.1 & 97.2 & 6.9 & 5.6\\
    3Mformer~\cite{wang20233mformer}& 94.8 & 98.7 & 92.0 & 93.8 & \textbf{97.8} & 58.5 & 6.7 \\
    ProtoGCN~\cite{liu2025protogcn} & 93.8 & 97.8 & 90.9 & 92.2& - & 43.4 & 24.9 \\
    \midrule
    SkeleT~\cite{yang2024expressive} & \textbf{97.0} & \textbf{99.6} & 94.6 & \textbf{96.4}& \underline{97.6} & 9.6 & 5.2\\
    BHaRNet-B$\dagger$~\cite{cho2025body} & 96.1 & 98.7 & 94.0 & 94.9 & 94.6 & \textbf{3.3} & \textbf{2.8}\\
    BHaRNet-E$\dagger$~\cite{cho2025body} & 96.2 & 98.8 & 94.3 & 95.0 & 94.6 & 5.4 & 5.5\\
    BHaRNet-P$\dagger$~\cite{cho2025body} & 96.3 & 98.8 & 94.3 & 95.2 & 95.3 & \underline{3.4} & 4.9\\
    \midrule
    \midrule
    \textbf{Ours (BHaRNet-B)} & 96.6 & 99.1 & 94.5 & 95.5 & 95.9 & 6.6 & 5.5\\
    \textbf{Ours (BHaRNet-E)} & 96.7 & \underline{99.2} & \underline{94.7} & 95.7 & 96.3 & 10.9 & 11.0\\
    \textbf{Ours (BHaRNet-P)} & \underline{96.8} & \underline{99.2} & \textbf{94.8} & \underline{95.8} & 95.9 & 6.8 & 9.7 \\
    
    \bottomrule
    \end{tabular}
    }
\end{table*}
\paragraph{Intra-skeleton Ensemble.}
Within the skeleton domain, four modalities—Joint (J), Bone (B), Joint Motion (JM), and Bone Motion (BM)—are aggregated as:
\begin{equation}
    \hat{y}_{\text{skel}} = 
    2(\hat{y}_J+\hat{y}_B) + (\hat{y}_{JM}+\hat{y}_{BM}),
\end{equation}
where spatial modalities receive higher weights to preserve structural stability, 
while motion modalities contribute complementary temporal sensitivity.

\paragraph{Skeleton–RGB Ensemble.}
In BHaRNet-M(Fig.~\ref{fig:bharnet_m}), the RGB stream follows the MMNet-based modulation strategy~\cite{bruce2022mmnet}, 
where skeleton features guide attention to motion-relevant regions.
The final prediction extends the deterministic ensemble to include appearance cues:
\begin{equation}
    \hat{y}_{\text{final}} =
    2(\hat{y}_J+\hat{y}_B) + (\hat{y}_{JM}+\hat{y}_{BM}) + 3\hat{y}_{\text{RGB}}.
\end{equation}
This five-modality integration (J, B, JM, BM, RGB) enriches both spatial–temporal and visual representations under a unified deterministic rule.
The fixed weights (2 : 2 : 1 : 1 : 3) were selected via grid search on a validation split.

\paragraph{Effect.}
Adding temporal modalities consistently improves recognition of fine-grained and hand-centric actions, while maintaining inference simplicity.
The fixed-weight ensemble reduces variance among modalities,
providing reliable performance under diverse noise and viewpoint conditions and further showing that the reliability-aware skeleton features remain beneficial when integrated with heterogeneous modalities.

\begin{figure}
    \centering
    \includegraphics[width=0.7\linewidth]{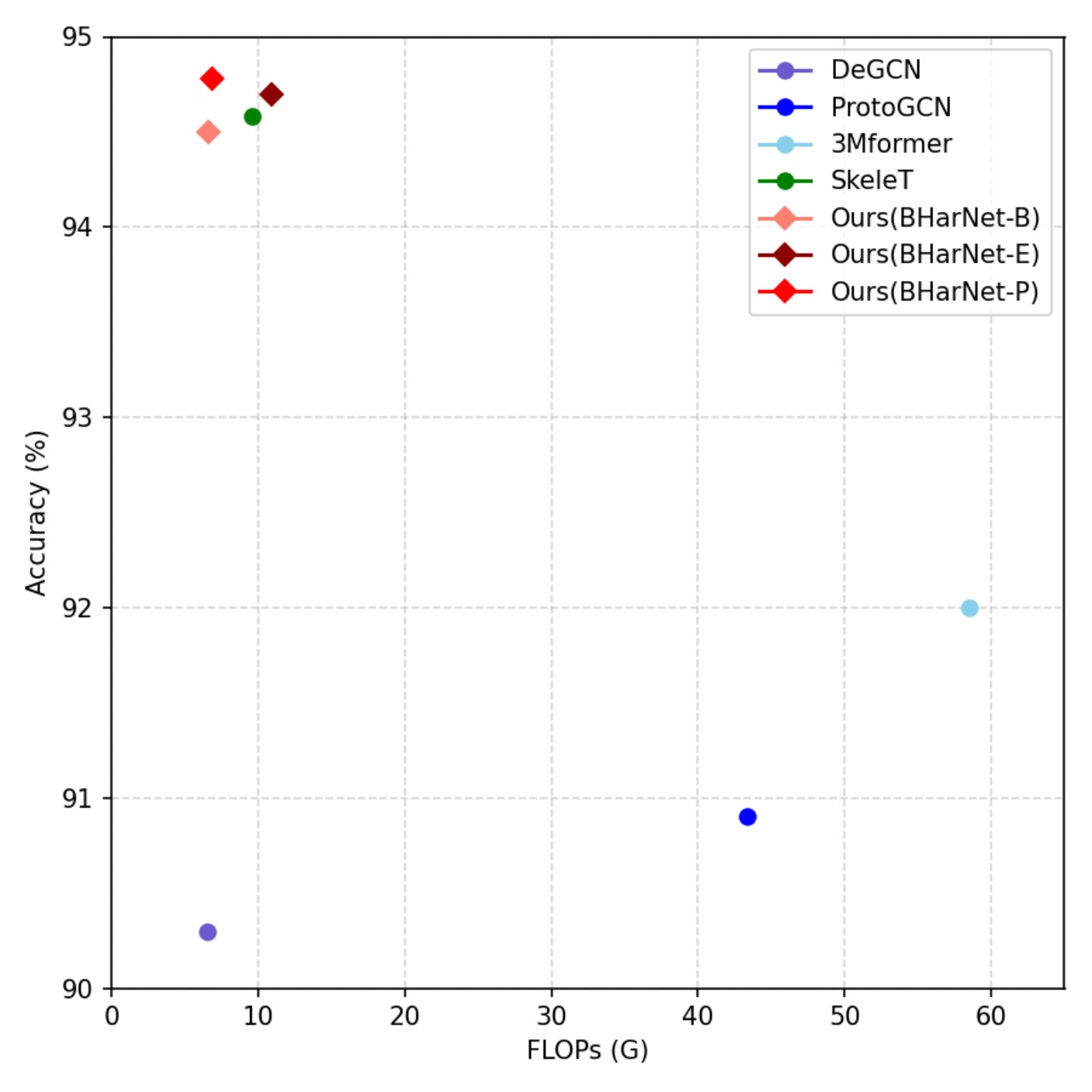}
    \caption{Accuracy–GFLOPs trade-off for skeleton-based action recognition on NTU 120 cross-subject.
    Red-toned markers denote our probabilistic models (BHaRNet-B/E/P), and green-to-blue markers denote previous skeleton-based state-of-the-art methods
    (DeGCN, ProtoGCN, 3MFormer, SkeleT).}
    \label{fig:plot_skel}
\end{figure}
\section{Experiments}
\label{sec:experiments}
\begin{table*}
    \centering
    \caption{Multi-modal action recognition on NTU RGB+D 60/120 and PKU-MMD benchmarks. “*” denotes experimental results provided in the referenced paper. Best and second-best results are highlighted in \textbf{bold} and \underline{underline}, respectively. BHaRNet-M$\dagger$ denotes the conference baseline, and “Ours” the proposed probabilistic variants.}
    \label{tab:ntu_multimodal}
    \resizebox{0.95\linewidth}{!}{
    \begin{tabular}{lccccccccc}
    \toprule
    \multirow{2}{*}{Method} &
    \multirow{2}{*}{Modality} &
    \multicolumn{2}{c}{NTU 60} &
    \multicolumn{2}{c}{NTU 120} &
    \multicolumn{2}{c}{PKU-MMD} &
    \multirow{2}{*}{GFLOPs} &
    \multirow{2}{*}{Params(M)} \\
    \cmidrule(lr){3-4}\cmidrule(lr){5-6}\cmidrule(lr){7-8}
     & & X-Sub & X-View & X-Sub & X-Set & X-Sub & X-view& & \\
    \midrule
    MMNet\cite{bruce2022mmnet} & J+B+RGB & 96.6 & 99.1 & 92.9 & 94.4 & \underline{97.4} & \underline{98.6} & 89.2 & 34.2 \\
    PoseConv3D\cite{duan2022poseconv}  & B+RGB & \textbf{97.0} & \textbf{99.6} & \underline{95.3} & \underline{96.4} & - & - & 41.8* & 31.6* \\
    $\pi$-ViT\cite{reilly2024justvit} & J+RGB & 94.0 & 97.9 & 91.9 & 92.9 & -& - & 590.0 & 121.4 \\
    EPAM-Net\cite{abdelkawy2025epam} & J+RGB & 96.1 & 99.0 & 92.4 & 94.3 & 96.2 & 98.4 & \underline{8.1} & \textbf{2.5} \\
    BHaRNet-M$\dagger$~\cite{cho2025body} & J+B+RGB & 96.3 & 99.0 & 95.1 & 96.0 & 96.9 & 97.9 & \textbf{7.6} & \underline{16.7} \\
    \midrule
    \midrule
    \multirow{2}{*}{\textbf{Ours (BHaRNet-M)}} & J+B+RGB & 96.8 & 99.3 & \underline{95.3} & \underline{96.2} & 97.3 & 98.5 & \textbf{7.6} & \underline{16.7} \\
    & J+B+JM+BM+RGB & \textbf{97.0} & \underline{99.4} & \textbf{95.5} & \textbf{96.5} & \textbf{97.5} & \textbf{98.7} & 13.0 & 22.2 \\
    \bottomrule
    \end{tabular}}
\end{table*}
We evaluate the proposed probabilistic dual-stream framework on major skeleton-based action recognition benchmarks, including NTU RGB+D 60/120~\cite{shahroudy2016ntu}, PKU-MMD~\cite{liu2017pku-mmd}, and N-UCLA~\cite{wang2014nucla}. 
We additionally define a new hand-centric benchmark, NTU-Hand~11/27, derived from hand-dominant classes within NTU 60/120, to evaluate fine-grained actions involving subtle hand articulations.  
We report two baseline references for clarity:
(i) BHaRNet$^\dagger$: the conference version as published, used as the official comparison baseline in all main result tables;
(ii) BHaRNet$^\ddagger$: a reproduced version trained under our updated calibration-free preprocessing. In the ablation studies (Section~\ref{sec:exp_ablation}), we treat BHaRNet$^\ddagger$ as the ``+Calibration-free'' step, and provide full results in Appendix~\ref{appendix:results}.

\subsection{Implementation Details}
\label{sec:exp_impl}
The training configuration follows the conference version~\cite{cho2025body} unless noted.  
For the skeleton model (BHaRNet-B/E/P), we adopt a two-stage training scheme: 
(1) pretraining the body and hand streams (DeGCN~\cite{myung2024degcn} backbone) separately, and 
(2) fine-tuning the full dual-stream model using the pretrained weights.  
For the multi-modal model (BHaRNet-M), an additional stage is introduced to train the RGB stream. 
All experiments are conducted on 1 RTX 3090 GPU.
Skeleton preprocessing steps follow the calibration-free design in Section~\ref{sec:method:canonical} and variant naming is provided in Appendix~\ref{appendix:implementation}.

\subsection{Main Results on Skeleton-based Recognition}
\label{sec:exp_main}
Table~\ref{tab:exp_skel} summarizes comparisons with state-of-the-art (SOTA) skeleton-based models on NTU RGB+D 60/120 and N-UCLA.  
Our method achieves SOTA accuracy on NTU 120 X-Sub and near-SOTA results on all remaining protocols, with competitive computational cost (6.6–10.9 GFLOPs).
While performance on N-UCLA is relatively modest, this dataset contains fewer hand-centric actions and limited training samples.
Nevertheless, our framework still improves noticeably over the conference version, indicating stronger generalization in low-data settings.

\noindent\textbf{Observations.}  
Compared to large-scale previous SOTA models (3MFormer~\cite{wang20233mformer}, SkeleT~\cite{yang2024expressive}), 
our framework achieves near-SOTA performance with roughly 30–90\% fewer FLOPs, maintaining strong efficiency–accuracy trade-offs (Fig.~\ref{fig:plot_skel}).
\subsection{Multi-modal Recognition with RGB Integration}
\label{sec:exp_multimodal}
Table~\ref{tab:ntu_multimodal} compares our multi-modal framework (BHaRNet-M) with recent pose–RGB approaches.  
The proposed model achieves state-of-the-art accuracy across NTU 120 and PKU-MMD, and highly competitive results on NTU 60, while preserving low computation.  
Under identical modality settings (J+B+RGB), our updated BHaRNet-M already surpasses its conference counterpart (BHaRNet-M$^\dagger$), demonstrating that the probabilistic learning and calibration-free design alone yield consistent gains by 0.2–0.6 percentage points (pp).  
Extending intra-skeleton motion cues (Joint Motion, Bone Motion) to RGB further improves by a small yet consistent margin (by 0.1–0.3 pp), suggesting that motion cues transfer effectively to RGB via body-guided modulation, reinforcing complementary temporal–appearance interactions.
\begin{figure}
    \centering
    \includegraphics[width=0.7\linewidth]{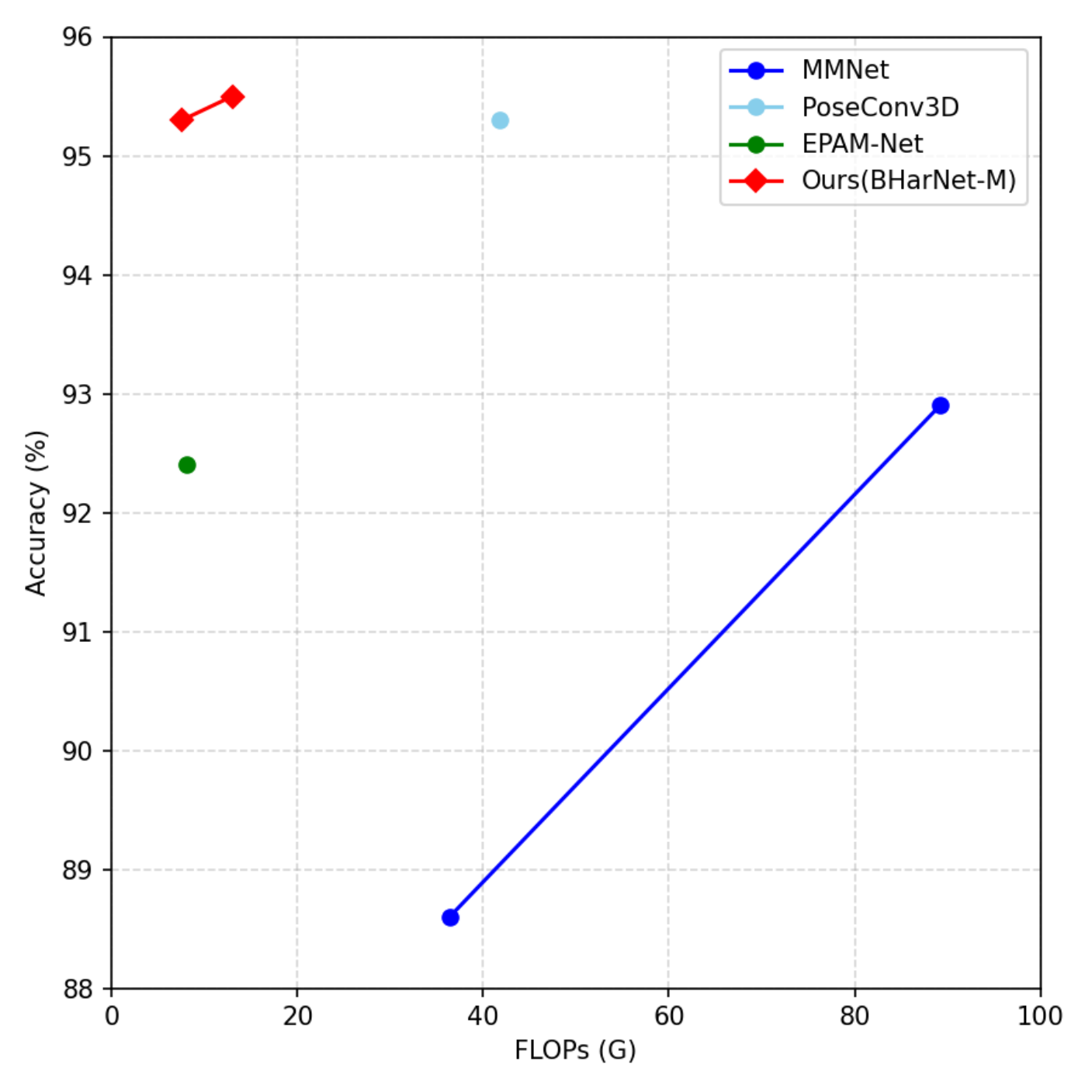}
    \caption{
    Accuracy–GFLOPs trade-off for multi-modal (skeleton+RGB) action recognition on NTU 120 cross-subject.
    Red markers denote our multi-modal framework (BHaRNet-M, left:J+B+RGB / right:J+B+JM+BM+RGB),
    and green-to-blue markers denote previous pose–RGB models
    (MMNet, PoseConv3D, EPAM-Net).
    }
    \label{fig:plot_rgb}
\end{figure}

\noindent\textbf{Efficiency.}  
Fig.~\ref{fig:plot_rgb} shows that, among pose–RGB methods, our five-modality BHaRNet-M (J+B+JM+BM+RGB) attains the highest NTU 120 X-Sub accuracy while using substantially fewer FLOPs than previous high-accuracy models. 
The model scales gracefully without noticeable overfitting, validating the stability of the probabilistic formulation across heterogeneous inputs.

\subsection{Hand-centric Benchmark: NTU-Hand 11/27}
\label{sec:exp_handcentric}
To evaluate fine-grained actions, we introduce NTU-Hand~11/27, consisting of 11 hand-centric classes from NTU 60 and additional 16 classes from NTU 120.
Hand-centric subsets contain actions with subtle hand articulations that contribute minimally in full-class evaluations.
Evaluation follows the standard X-Sub/X-View (NTU-Hand 11) and X-Sub/X-Set (NTU-Hand 27) protocols.  
Table~\ref{tab:exp_hand} shows that our method achieves the best or on-par performance on both splits, 
outperforming SkeleT (up to 2 pp) and ProtoGCN (by 4–9 pp) on hand-centric subsets.  
These results highlight that reliability modeling effectively captures fine-scale articulations often overlooked in body-centric models.
Detailed class lists, per-class accuracy, and full variant comparisons are provided in Appendix~\ref{appendix:results}.

\begin{table*}
    \centering
    \caption{Performance Comparison with State-of-the-Art Methods for Skeleton-based Action Recognition on NTU-Hand 11/27. “NTU-Hand 11” refers to 11 hand-centric classes within 60 classes of NTU RGB+D 60, and “NTU-Hand 27” to 27 classes within 120 classes of NTU RGB+D 120.}
    \label{tab:exp_hand}
    \resizebox{0.5\linewidth}{!}{
    \begin{tabular}{lcccc}
    \toprule
    \multirow{2}{*}{Method} & 
    \multicolumn{2}{c}{NTU-Hand 11} &
    \multicolumn{2}{c}{NTU-Hand 27} \\ 
    \cmidrule(lr){2-3}\cmidrule(lr){4-5}
    
     & X-Sub & X-View & X-Sub & X-Set\\
    \midrule
    ProtoGCN~\cite{liu2025protogcn} & 87.4 & 94.4 & 82.0 & 84.7\\
    \midrule
    SkeleT~\cite{yang2024expressive} & 94.7 & \textbf{98.6} & 89.9 & \underline{92.6}\\
    BHaRNet-P$\dagger$ & 95.4 & 97.8 & 90.8 & 92.2\\
    \midrule
    \midrule
    \textbf{Ours (BHaRNet-E)} & \underline{95.7} & \underline{98.5} & \underline{91.4} & \textbf{93.1} \\
    \textbf{Ours (BHaRNet-P)} & \textbf{96.1} & \underline{98.5} & \textbf{91.7} & \textbf{93.1} \\ 
    \bottomrule
    \end{tabular}
    }
\end{table*}

\subsection{Ablation Studies}
\label{sec:exp_ablation}
We conduct a series of ablations to quantify the contribution of each component.
We evaluate three progressively enhanced configurations:
(i) Baseline$\dagger$, (ii) +Calibration-free preprocessing (equivalent to the reproduced baseline, BHaRNet$\ddagger$),
(iii) +Noisy-OR loss (our full probabilistic model).
Importantly, these steps introduce no additional parameters or FLOPs, isolating the effect of reliability-aware learning without conflating performance gains with architectural expansion.
Comprehensive results, including variant-level breakdowns and modality extensions across datasets, are provided in Appendix~\ref{appendix:results}, and show consistent trends with the improvements reported in the main tables.

\begin{table*}
    \centering
    \caption{Ablation Study on modality extension on NTU RGB+D 120 and NTU-Hand 27 cross-subject.}
    \label{tab:abl_extension}
    \resizebox{0.6\linewidth}{!}{
    \begin{tabular}{lccc}
    \toprule
    Method & Modality & NTU 120 & NTU-Hand 27 \\
    \midrule
    \multirow{2}{*}{\textbf{Ours (BHaRNet-E)}} & J+B & 94.5& 91.3\\
    & J+B+JM+BM & 94.7 & 91.4\\
    \textbf{Ours (BHaRNet-M)} & J+B+JM+BM+RGB & \textbf{95.5} &  \textbf{92.4} \\
    \bottomrule
    \end{tabular}
    }
\end{table*}

\subsubsection{Modality Extension.}
Table~\ref{tab:abl_extension} reports step-by-step improvements from the modality extension modules. 
Starting from our probabilistic J+B configuration, we further add JM/BM and RGB.
Notably, these improvements appear consistently across NTU 120 and NTU-Hand 27, indicating that the probabilistic formulation benefits both large-scale and fine-grained actions. 
Integrating temporal motion cues (JM, BM) yields further enhancement, and adding RGB achieves the highest accuracy, demonstrating that each component contributes complementary robustness.

\subsubsection{Cross-View Robustness.}
Table~\ref{tab:abl_xview} evaluates performance under view variation.  
Our updated framework consistently improves cross-view generalization, suggesting that canonical-space alignment in the conference version may have overfit to specific camera geometries.

\begin{table}
    \centering
    \caption{Ablation Study for cross-viewpoint evaluation on NTU RGB+D 60 and NTU-Hand 11. Comparison for Joint-Bone ensembled model.}
    \label{tab:abl_xview}
    \resizebox{0.85\linewidth}{!}{
    \begin{tabular}{lccc}
    \toprule
    \multirow{2}{*}{Method} 
    & NTU 60 & NTU-Hand 11 \\
    \cmidrule(lr){2-2}\cmidrule(lr){3-3}
    & X-View & X-View \\
    \midrule
    BHaRNet-E$\dagger$ & 98.8 & 97.9\\
    \textbf{Ours (BHaRNet-E)} & \textbf{99.0} & \textbf{98.2}\\
    \bottomrule
    \end{tabular}
    }
\end{table}

\subsubsection{Novelty Breakdown.}
Table~\ref{tab:ablation_complementary} decomposes the effect of calibration-free preprocessing and Noisy-OR loss for each variant (B, E, P).  
All variants exhibit consistent improvements across NTU 120 and NTU-Hand 27, confirming that the proposed modules generalize across variant styles.
The only exception is BHaRNet-B on NTU-Hand~27, where the calibration-free variant slightly outperforms the additional Noisy-OR regularization.
We attribute this to a conservative shift toward body cues in the expert-only setting and provide a detailed analysis in our robustness study (Section~\ref{sec:exp_robust}).

\begin{table}
    \centering
    \caption{Ablation study for novelty steps in NTU 120 and NTU-Hand 27 cross-subject benchmarks. Processed in Joint modality.}
    \label{tab:ablation_complementary}
    \resizebox{0.8\linewidth}{!}{
    \begin{tabular}{lcc}
    \toprule
    Method & NTU 120 & NTU-Hand 27 \\
    \midrule
    BHaRNet-B$\dagger$ & 92.7 & 88.6 \\
    + Calibration-free & 93.0 & \textbf{89.4} \\
    + Noisy-OR Loss & \textbf{93.2} & 89.2 \\
    \midrule
    BHaRNet-E$\dagger$ & 93.0 & 88.8 \\
    + Calibration-free & 93.3 & 89.4 \\
    + Noisy-OR Loss & \textbf{93.5} & \textbf{89.5} \\
    \midrule
    BHaRNet-P$\dagger$ & 92.9 & 88.7 \\
    + Calibration-free & 93.3 & 89.4\\
    + Noisy-OR Loss & \textbf{93.4} & \textbf{89.5} \\
    \bottomrule
    \end{tabular}}
\end{table}

\begin{table}[t]
    \centering
    \caption{Frame-drop robustness on NTU RGB+D 120 and NTU-Hand 27 cross-subject benchmarks. We report accuracy under different frame missing rates (0, 0.25, 0.5).}
    \label{tab:abl_noise}
    \resizebox{0.95\linewidth}{!}{
    \begin{tabular}{lcccccc}
    \toprule
    \multirow{2}{*}{Method} & 
    \multicolumn{3}{c}{NTU 120} &
    \multicolumn{3}{c}{NTU-Hand 27} \\ 
    \cmidrule(lr){2-4}\cmidrule(lr){5-7}
    
     & 0 & 0.25 & 0.5 & 0 & 0.25 & 0.5 \\
    \midrule
    BHaRNet-B$\dagger$ & 92.7 & 90.1 & 89.8 & 88.6 & 88.6 & 88.0\\
    + Calibration-free & 93.0 & 92.8 & 92.6 & \textbf{89.4} & 88.7 & 88.4\\
    + Noisy-OR Loss & \textbf{93.2} & \textbf{93.0} & \textbf{92.8} & 89.2 & \textbf{89.1} & \textbf{88.6}\\
    \midrule
    BHaRNet-E$\dagger$ & 93.0 & 90.1 & 90.1 & 88.8 & 88.9 & 88.3\\
    + Calibration-free & 93.3 & 93.2 & 93.0 & 89.4 & 88.8 & 88.9\\
    + Noisy-OR Loss & \textbf{93.5} & \textbf{93.3} & \textbf{93.1} & \textbf{89.5} & \textbf{89.3} & \textbf{89.1} \\
    \bottomrule
    \end{tabular}
    }
\end{table}
\begin{figure}
    \centering
    \begin{minipage}[t]{0.98\columnwidth}
        \centering
        \includegraphics[width=\linewidth]{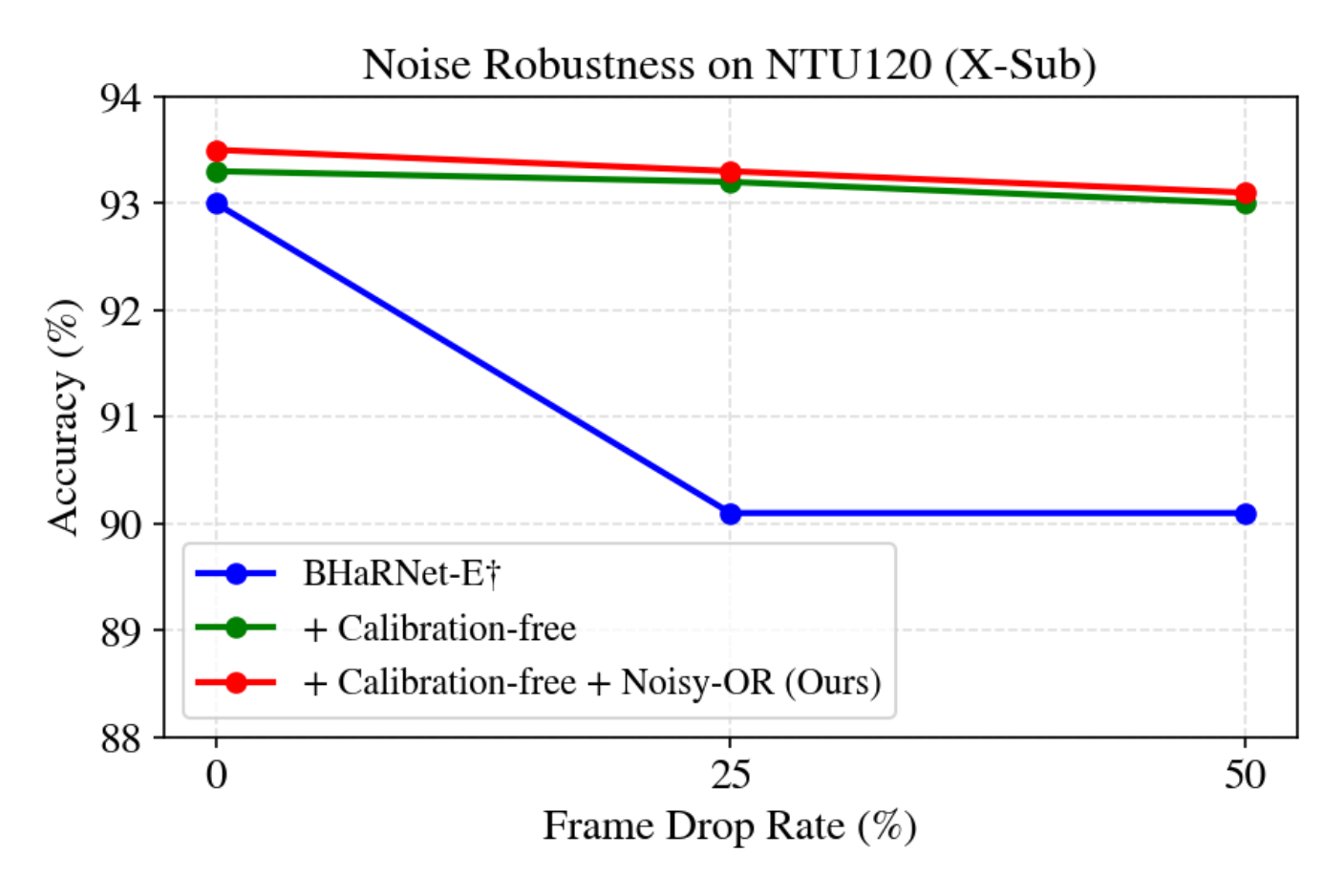} 
    \end{minipage}
    \hfill
    \begin{minipage}[t]{0.98\columnwidth}
        \centering
        \includegraphics[width=\linewidth]{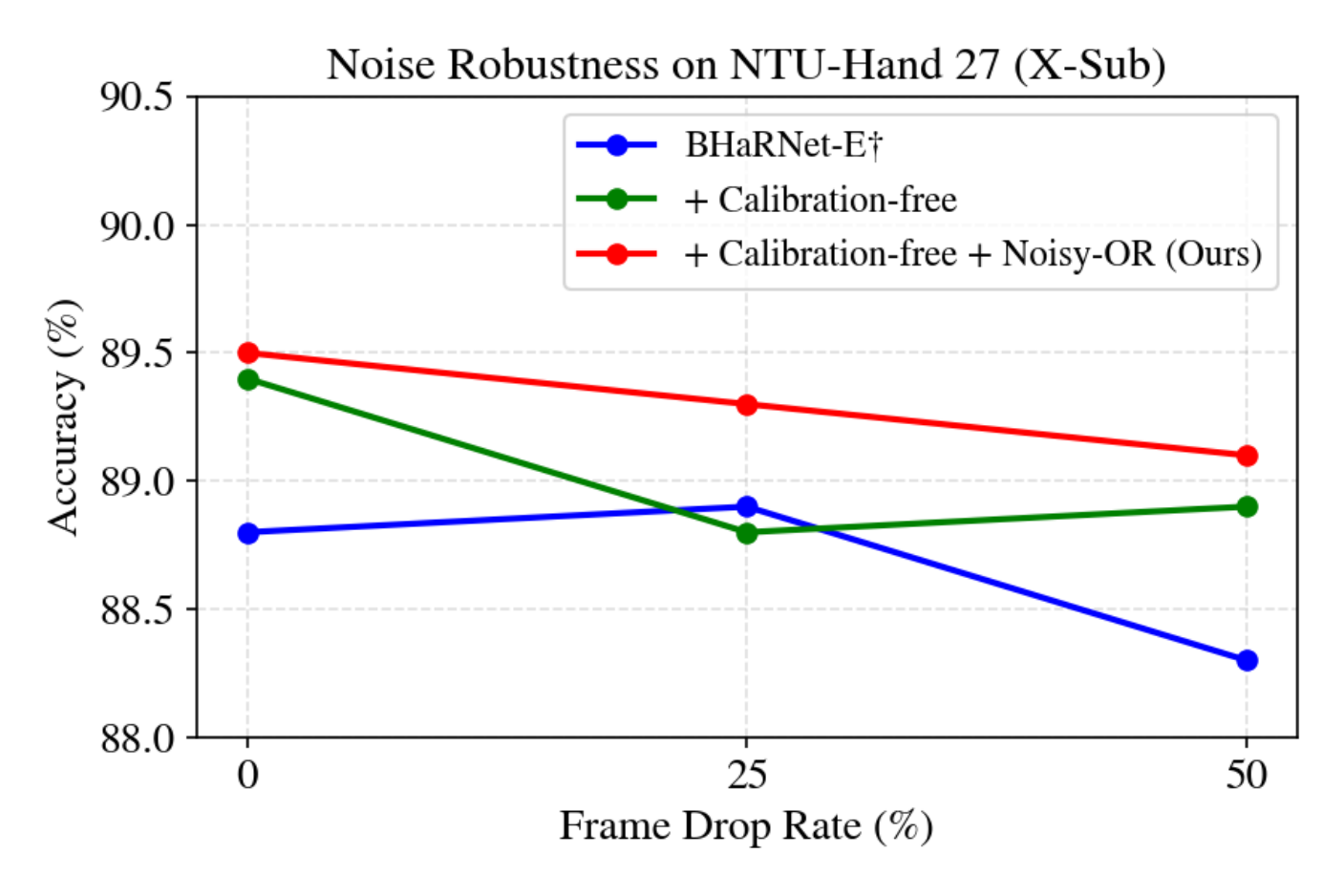} 
    \end{minipage}
    \caption{
    Noise robustness under frame-drop conditions on cross-subject benchmarks of NTU 120(top) and NTU-Hand 27(bottom) for BHaRNet-E models.
    We evaluate the conference baseline BHaRNet-E$^\dagger$ (blue line) and our probabilistic model 
    with calibration-free preprocessing(green line) and with calibration-free plus Noisy-OR loss (red line)
    at frame missing rates of 0, 25, and 50\%.
    }
    \label{fig:plot_noise}
\end{figure}

\begin{figure}
    \centering
    \includegraphics[width=0.8\linewidth]{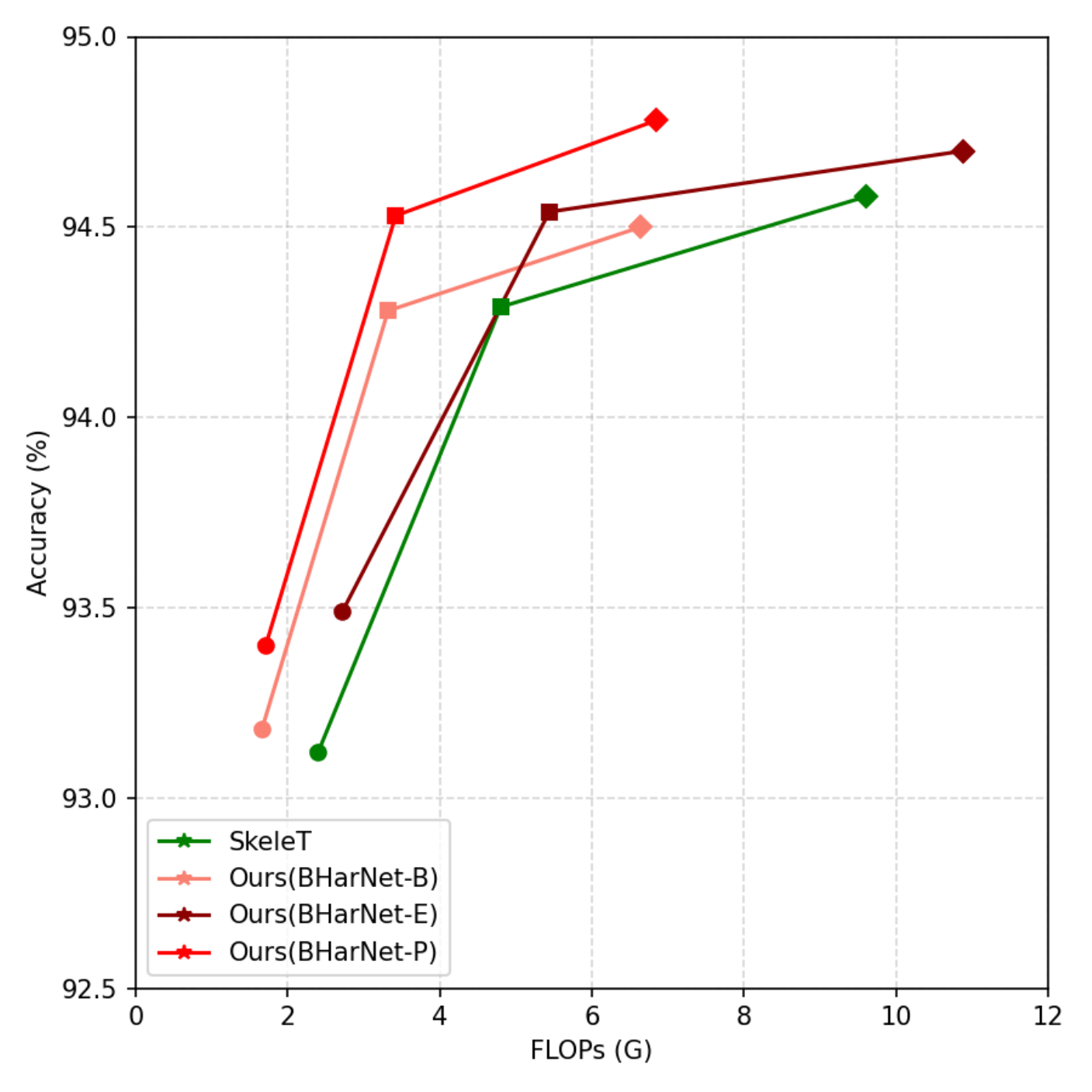}
    \caption{
    Accuracy–GFLOPs comparison in NTU 120 cross-subject benchmark between SkeleT and our probabilistic framework across 1-modality(left of line, marked as $\CIRCLE$), 2-modality(middle, $\blacksquare$), and 4-modality(right, $\blacklozenge$) configurations.
    Green-lined markers denote SkeleT variants, and red-toned-lined markers denote our corresponding BHaRNet variant(B/E/P).
    }
    \label{fig:plot_efficiency}
\end{figure}
\subsubsection{Noise Robustness.}
\label{sec:exp_robust}
Table~\ref{tab:abl_noise} analyzes robustness under frame-drop conditions (0, 0.25, 0.5) for both expert-only (BHaRNet-B) and expert+interactive (BHaRNet-E) configurations. 
For the canonical-space baselines (BHaRNet-B$^\dagger$/E$^\dagger$), accuracy degrades sharply as the frame-drop rate increases, with drops of more than 2--3 pp on NTU 120.
Removing the canonical transform already yields much flatter curves, especially for BHaRNet-E, indicating that calibration-free preprocessing mitigates the amplification of temporal noise due to corrupted hand frames.
Interestingly, the baseline exhibits a slight improvement under 25\% frame-drop. This occurs because canonical-space alignment amplifies hand-joint noise; removing a portion of heavily corrupted frames effectively reduces this propagated noise, resulting in a mild denoising effect. In contrast, our probabilistic formulation already suppresses unreliable cues, making frame removal unnecessary and yielding stable performance (Fig.~\ref{fig:plot_noise}).

With the Noisy-OR loss, our models also become robust on NTU-Hand 27 as well, showing stable reliability-aware fusion.
For both BHaRNet-B and BHaRNet-E, the gap between 0\% and 50\% frame-drop shrinks in a smooth curve on NTU-Hand~27, whereas calibration-free baselines suffer substantially larger degradation.
Also for the expert-only variant BHaRNet-B, adding the Noisy-OR loss yields more stable accuracy under frame-drop perturbations but slightly reduces hand-centric performance compared to the calibration-free baseline.
We hypothesize that, without interactive branches, the probabilistic regularizer biases the expert
towards more conservative body cues, which is compensated in the full BHaRNet-E configuration
where interactive branches can recover fine-grained hand information.

\subsection{Efficiency and Trade-off Analysis}
Across skeleton-only and multi-modal settings, our framework achieves higher or comparable accuracy at similar or lower FLOPs than most prior methods, with substantial gains over body-centric models.
Fig.~\ref{fig:plot_efficiency} further visualizes the accuracy–GFLOPs trade-off against SkeleT across 1/2/4-modality settings (J/J+B/J+B+JM+BM), indicating that our models lie on or near a more favorable accuracy–efficiency frontier than SkeleT.
Grid-search on validation data shows that fixed ensemble weights (2:2:1:1:3) remain stable within $\pm50\%$ perturbation ($<$0.1 pp difference), 
supporting parameter-free inference and robustness across configurations.
These results verify that the proposed formulation improves robustness without increasing inference complexity, aligning with the design goal of reliability-aware modeling.

\section{Conclusion}
\label{sec:conclusion}

We presented a generalized probabilistic dual-stream framework for body--hand action recognition that unifies calibration-free skeleton learning, reliability-aware fusion, and multi-modal integration. 
Extensive experiments show that the proposed framework achieves state-of-the-art or competitive performance at similar computational cost, consistently outperforming our conference baseline. 
The new hand-centric benchmark (NTU-Hand 11/27) further highlights clear gains on fine-grained, hand-dominant actions. 
Ablation studies confirm that calibration-free learning and the Noisy-OR loss provide complementary benefits, improving robustness to viewpoint changes and frame-drop perturbations without increasing inference complexity.

Despite these advantages, our framework is ultimately bounded by the quality of the underlying motion signals and still shows weakness on small datasets such as N-UCLA.
In third-person HAR videos, hands are often small, occluded, or blurred, making the extracted hand skeletons inherently noisy and sometimes unreliable. 
In this work, we deliberately focus on a single RGB-based skeleton extraction pipeline on standard benchmarks to isolate the effect of the proposed reliability-aware learning.
Extending the framework to diverse sensing setups—such as RGB-D cameras, wearable or egocentric sensors—and studying how sensing choices interact with reliability modeling is an interesting direction for future work.
\section{Acknowledgement}
We would like to thank Hye-Yeon Kim for her valuable help and support throughout this work.

\appendix
\setcounter{table}{0}
\setcounter{figure}{0}
\setcounter{equation}{0}
\renewcommand{\thetable}{\Alph{section}.\arabic{table}}
\renewcommand{\thefigure}{\Alph{section}.\arabic{figure}}
\renewcommand{\theequation}{\Alph{section}.\arabic{equation}}

\section{Preprocessing and Hand-Centric Benchmark Details}
\label{appendix:preprocess}

\subsection{Body--Hand Skeleton Construction}

We follow the body--hand preprocessing pipeline described in Section~\ref{sec:method:canonical}.
Given RGB video frames, body and hand keypoints are estimated using MediaPipe~\cite{lugaresi2019mediapipe}. 
We then construct a body skeleton sequence $X_B \in \mathbb{R}^{C \times T \times V_B}$ and a hand skeleton sequence
$X_H \in \mathbb{R}^{C \times T \times V_H}$, where $C$ denotes the coordinate channels $(x,y,z)$, $T$ the temporal length, 
and $V_B, V_H$ the numbers of body and hand joints, respectively.

\paragraph{Temporal processing.}
We first identify valid frames with reliable hand detections and apply temporal smoothing by filtering these frames and linearly interpolating across short gaps.
The resulting sequences are then resampled to a fixed length and, when necessary, padded by stacking boundary frames, following common practice in recent works.

\paragraph{Spatial processing.}
For the hand stream, we use the 21-joint Mediapipe hand graph (including its native kinematic edges).
For datasets with 25 body joints (NTU RGB+D 60/120, PKU-MMD), we attach the 21 hand joints and reserve 4 dummy joints with zero-masked coordinates to maintain consistent topology.
For N-UCLA, which provides 20 body joints, we remove the THUMB\_IP joint from the hand to match the reduced body joint configuration.

\paragraph{Hand-centric strategy.}
In our conference framework, the hand center node was defined with respect to the most active one in the scene 
and used for relative coordinate normalization, as body preprocessing does.
In contrast, for hand modeling we define a separate local center for each hand using its wrist joint,
so that large global body motion does not dominate local hand motion. This design reduces apparent motion blur in the hand stream while preserving the shared camera coordinate system used in the main model.

\subsection{NTU-Hand 11/27 Class Definitions}

The NTU-Hand benchmark is constructed as a hand-centric subset of NTU RGB+D 60/120~\cite{shahroudy2016ntu}. 
We select actions in which subtle hand articulations play a dominant role, and define two subsets:

\paragraph{NTU-Hand 11.}
This subset consists of 11 hand-centric classes from NTU RGB+D 60:
\textit{clapping, reading, writing, tear up paper, phone call, play with phone/tablet, type on a keyboard, point to something, taking a selfie, check time (from watch), rub two hands}.

\paragraph{NTU-Hand 27.}
This subset extends NTU-Hand 11 by adding 16 hand-centric classes from NTU RGB+D 120:
\textit{thumb up, thumb down, make OK sign, make victory sign, staple book, counting money, cutting nails, cutting paper, snap fingers, open bottle, sniff/smell, squat down, toss a coin, fold paper, ball up paper, play magic cube}.
Both NTU-Hand 11 and NTU-Hand 27 are evaluated under the original NTU RGB+D 60/120 cross-subject and cross-view/setup protocols,
restricted to the selected hand-centric classes.

\section{Implementation Details}
\label{appendix:implementation}

\subsection{Network Variants and Branch Configuration}

Our dual-stream architecture builds on DeGCN backbones~\cite{myung2024degcn} for both body and hand streams. 
We define three skeleton-only variants and one multi-modal variant:
\begin{itemize}
    \item \textbf{BHaRNet-B}: An expert-only configuration with body and hand expert branches (BE, HE).   
    \item \textbf{BHaRNet-E}: An expert+interactive configuration, combining expert branches (BE, HE) with interactive body/hand branches (BI, HI) to share context.
    \item \textbf{BHaRNet-P}: A parameter-efficient configuration using lightweight interactive branches with reduced channel width. 
    \item \textbf{BHaRNet-M}: A multi-modal variant that extends BHaRNet-E with an RGB stream and MMNet-style body-guided modulation, as described in Section~\ref{sec:background:baseline}.
\end{itemize}

\subsection{Training Protocol}

Unless otherwise noted, we follow the training settings of DeGCN and our conference framework~\cite{cho2025body}. 
For the skeleton models (BHaRNet-B/E/P), we adopt a two-stage procedure: 
(1) pretrain the body and hand streams independently on their respective skeleton inputs, and 
(2) fine-tune the full dual-stream architecture with all losses enabled. 
The loss weights $(\lambda_{\text{idv}}, \lambda_{\text{cpl}}, \lambda_{\text{nor}})$ are fixed across all second stage experiments, and 
inference uses deterministic logit summation without any stochastic component or test-time augmentation. 

After processing steps for 4 intra-skeleton modalities, we introduce a third stage for the multi-modal model (BHaRNet-M). We train RGB stream with skeleton-guided modulation while freezing the pretrained skeleton backbones.

\section{Additional Quantitative Results}
\label{appendix:results}

This section provides extended quantitative results that complement the analyses in Section~\ref{sec:experiments}. 
Table~\ref{tab:appendix:ablation_modality} aggregates and extends the modality and variant ablations 
(Tables~\ref{tab:exp_hand},~\ref{tab:abl_extension} and~\ref{tab:abl_xview}). 
Table~\ref{tab:appendix:ablation_framedrop} jointly expands the analysis of frame-drop robustness and novelty breakdown 
(Tables~\ref{tab:ablation_complementary} and~\ref{tab:abl_noise}).
Table~\ref{tab:appendix:exp_skel} provides an extended comparison with skeleton-based baselines 
(Table~\ref{tab:exp_skel}). 
These generalized tables do not change the main conclusions, but offer a more complete view of the behavior of our probabilistic framework across datasets, modalities, and perturbation settings.

\begin{table*}[t]
    \centering
    \caption{Full ablation of baseline vs.\ probabilistic variants and modality configurations 
    (J+B, J+B+JM+BM, J+B+JM+BM+RGB) on NTU RGB+D 60/120 and NTU-Hand 11/27.}
    \label{tab:appendix:ablation_modality}
    \resizebox{0.95\linewidth}{!}{
    \begin{tabular}{lccccccccc}
    \toprule
    \multirow{2}{*}{Method} & 
    \multirow{2}{*}{Modality} & 
    \multicolumn{2}{c}{NTU 60} &
    \multicolumn{2}{c}{NTU 120} & 
    \multicolumn{2}{c}{NTU-Hand 11} &
    \multicolumn{2}{c}{NTU-Hand 27}\\ 
    \cmidrule(lr){3-4}\cmidrule(lr){5-6}\cmidrule(lr){7-8}\cmidrule(lr){9-10}
    
     & & X-Sub & X-View & X-Sub & X-Set &  X-Sub & X-View & X-Sub & X-Set \\
    \midrule
    BHaRNet-B† &J+B & 96.1 & 98.7 & 94.0 & 94.9 & 95.4 & 97.9 & 90.6 & 91.8 \\
    BHaRNet-E† &J+B & 96.2 & 98.8 & 94.3 & 95.0 & 95.6 & 97.9 & 90.9 & 92.0 \\
    BHaRNet-P† &J+B & 96.3 & 98.8 & 94.3 & 95.2 & 95.4 & 97.8 & 90.8 & 92.2\\
    \midrule
    \textbf{Ours (BHaRNet-B)} & J+B & 96.4 & 99.1 & 94.3 & 95.2 & 95.6 & 98.2 & 91.3 & 92.7\\
    \textbf{Ours (BHaRNet-E)} & J+B & 96.6 & 99.0 & 94.5 & 95.4 & 95.6 & 98.2 & 91.3 & 92.8\\
    \textbf{Ours (BHaRNet-P)} & J+B & 96.7 & 99.1 & 94.5 & 95.5 & 95.8 & 98.1 & 91.4 & 92.9\\
    \midrule
    \textbf{Ours (BHaRNet-B)} & J+B+JM+BM & 96.6 & 99.1 & 94.5 & 95.5 & 95.7 & 98.4 & 91.4 & 93.0\\
    \textbf{Ours (BHaRNet-E)} & J+B+JM+BM & 96.7 & 99.2 & 94.7 & 95.7 & 95.7 & 98.5 & 91.4 & 93.1\\
    \textbf{Ours (BHaRNet-P)} & J+B+JM+BM & 96.8 & 99.2 & 94.8 & 95.8 & 96.1 & 98.5 & 91.7 & 93.1\\
    \midrule
    \textbf{Ours (BHaRNet-M)} & J+B+JM+BM+RGB & 97.0 & 99.4 & 95.5 & 96.5 & 96.1 & 98.8 & 92.4 & 94.1\\
    \bottomrule
    \end{tabular}
    }
\end{table*}

\begin{table*}[t]
    \centering
    \caption{Full ablation of Frame-drop robustness for canonical, calibration-free, and Noisy-OR variants on 
    NTU RGB+D 120 and NTU-Hand 27 (cross-subject, Joint modality).}
    \label{tab:appendix:ablation_framedrop}
    \resizebox{0.5\linewidth}{!}{
    \begin{tabular}{lcccccc}
    \toprule
    \multirow{2}{*}{Method} & 
    \multicolumn{3}{c}{NTU 120} &
    \multicolumn{3}{c}{NTU-Hand 27} \\ 
    \cmidrule(lr){2-4}\cmidrule(lr){5-7}
    
     & 0 & 0.25 & 0.5 & 0 & 0.25 & 0.5 \\
    \midrule
    BHaRNet-B† & 92.7 & 90.1 & 89.8 & 88.6 & 88.6 & 88.0\\
    + Calibration-free & 93.0 & 92.8 & 92.6 & 89.4 & 88.7 & 88.4\\
    + Noisy-OR Loss & 93.2 & 93.0 & 92.8 & 89.2 & 89.1 & 88.6\\
    \midrule
    BHaRNet-E† & 93.0 & 90.1 & 90.1 & 88.8 & 88.9 & 88.3\\
    + Calibration-free & 93.3 & 93.2 & 93.0 & 89.4 & 88.8 & 88.9\\
    + Noisy-OR Loss & 93.5 & 93.3 & 93.1 & 89.5 & 89.3 & 89.1 \\
    \midrule
    BHaRNet-P† & 92.9 & 91.0 & 90.9 & 88.7 & 89.0 & 88.3\\
    + Calibration-free & 93.3 & 93.2 & 93.0 & 89.4 & 89.1 & 88.7\\
    + Noisy-OR Loss & 93.4 & 93.3 & 93.0 & 89.5 & 89.5 & 88.7\\  
    
    \bottomrule
    \end{tabular}
    }
\end{table*}

\begin{table*}[t]
    \centering
    \caption{Extended comparison of skeleton-based methods on NTU RGB+D 60/120 and N-UCLA:
    accuracy (\%), GFLOPs, and number of parameters. “–” indicates results not reported in the
    original reference, and “*” denotes results reproduced using public code.}
    \label{tab:appendix:exp_skel}
    \resizebox{0.8\linewidth}{!}{
    \begin{tabular}{lccccccc}
    \toprule
    \multirow{2}{*}{Method} & 
    \multicolumn{2}{c}{NTU 60} &
    \multicolumn{2}{c}{NTU 120} &
    {N-UCLA} &
    \multirow{2}{*}{GFLOPs} & 
    \multirow{2}{*}{Params(M)} \\ \cmidrule(lr){2-3}\cmidrule(lr){4-5}\cmidrule(lr){6-6}
    
     &  X-Sub & X-View & X-Sub & X-Set & X-View & \\
    \midrule
    CTR-GCN~\cite{chen2021ctrgcn} & 92.4 & 96.8 & 88.9 & 90.6 & 96.5 & 7.9 & 5.8 \\
    InfoGCN~\cite{chi2022infogcn} & 93.0 & 97.1 & 89.8 & 91.2 & 97.0 & 10.0* & 9.4\\
    PoseConv3D~\cite{duan2022poseconv} & 94.1 & 97.1 & 86.9 & 90.3 & - & 31.8 & \underline{4.0} \\
    BlockGCN~\cite{zhou2024blockgcn} & 93.1 & 97.0 & 90.3 & 91.5 & 96.9 & 6.5 & 5.2\\
    DeGCN~\cite{myung2024degcn} & 93.6 & 97.4 & 91.0 & 92.1 & 97.2 & 6.9 & 5.6\\
    3Mformer~\cite{wang20233mformer}& 94.8 & 98.7 & 92.0 & 93.8 & \textbf{97.8} & 58.5 & 6.7 \\
    ProtoGCN~\cite{liu2025protogcn} & 93.8 & 97.8 & 90.9 & 92.2& - & 43.4 & 24.9 \\
    \midrule
    SkeleT~\cite{yang2024expressive} & \textbf{97.0} & \textbf{99.6} & 94.6 & \textbf{96.4}& \underline{97.6} & 9.6 & 5.2\\
    BHaRNet-B$\dagger$~\cite{cho2025body} & 96.1 & 98.7 & 94.0 & 94.9 & 94.6 & \textbf{3.3} & \textbf{2.8}\\
    BHaRNet-E$\dagger$~\cite{cho2025body} & 96.2 & 98.8 & 94.3 & 95.0 & 94.6 & 5.4 & 5.5\\
    BHaRNet-P$\dagger$~\cite{cho2025body} & 96.3 & 98.8 & 94.3 & 95.2 & 95.3 & \underline{3.4} & 4.9\\
    \midrule
    BHaRNet-B$\ddagger$ &96.5&99.0&94.2&95.2&95.7& 9.6 & 5.2\\
    BHaRNet-E$\ddagger$ &96.6&99.0&94.4&95.5&95.5 & \textbf{3.3} & \textbf{2.8}\\
    BHaRNet-P$\ddagger$ &96.6&99.1&94.5&95.5&95.0& \underline{3.4} & 4.9\\
    \midrule
    \midrule
    \textbf{Ours (BHaRNet-B)} & 96.6 & 99.1 & 94.5 & 95.5 & 95.9 & 6.6 & 5.5\\
    \textbf{Ours (BHaRNet-E)} & 96.7 & \underline{99.2} & \underline{94.7} & 95.7 & 96.3 & 10.9 & 11.0\\
    \textbf{Ours (BHaRNet-P)} & \underline{96.8} & \underline{99.2} & \textbf{94.8} & \underline{95.8} & 95.9 & 6.8 & 9.7 \\
    
    \bottomrule
    \end{tabular}
    }
\end{table*}

\section{Additional Qualitative Visualization}
\label{appendix:vis}

We provide qualitative samples as supplementary video material (Video~S1). 
The video illustrates the behavior of our dual-stream body--hand framework, showing per-branch predictions (body and hand streams) and their ensemble together with the corresponding body and hand keypoint estimations on sample sequences from NTU  dataset.
Each keypoint is rendered as a node whose radius is scaled according to the local joint motion magnitude, so that joints with larger articulated motion appear more prominent in the visualization.

\printcredits

\bibliographystyle{cas-model2-names}

\bibliography{cas-refs}


\bio{figures/cho_bio}
\textbf{Seungyeon Cho} received the B.S. and M.S. degrees from the Korea Advanced Institute of Science and Technology (KAIST), Daejeon, Republic of Korea. 
His research interests include skeleton-based human action recognition, body–hand coordination, reliability-aware learning, efficient and multi-modal modeling, and further to the application of machine learning to photonics and inverse-designed optical devices.
\endbio

\bio{figures/kim_bio}
\textbf{Tae-Kyun (T-K) Kim} is a Professor and the director of Computer Vision and Learning Lab at School of Computing, KAIST, since 2020, and has been a Reader at Imperial College London, UK in 2010- 2024. He obtained his PhD from Univ. of Cambridge in 2008 and Junior Research Fellowship (governing body) of Sidney Sussex College, Univ. of Cambridge during 2007-2010. His BSc and MSc are from KAIST. His research interests lie in machine (deep) learning for 3D vision and generative AI, he has co-authored over 100 academic papers in top-tier conferences and journals in the field. He was the general chair of BMVC17 in London, and the program co-chair of BMVC23, Associate Editor of TPAMI, Pattern Recognition Journal, Image and Vision Computing Journal. He regularly serves as an Area Chair for the top-tier AI/vision conferences.
\endbio

\end{document}